\documentclass[lettersize,journal]{IEEEtran}
\usepackage{amsmath,amsfonts}
\usepackage{algorithmic}
\usepackage{array}
\usepackage[caption=false,font=normalsize,labelfont=sf,textfont=sf]{subfig}
\usepackage{textcomp}
\usepackage{stfloats}
\usepackage{url}
\usepackage{verbatim}
\usepackage{graphicx}
\def\BibTeX{{\rm B\kern-.05em{\sc i\kern-.025em b}\kern-.08em
    T\kern-.1667em\lower.7ex\hbox{E}\kern-.125emX}}
\usepackage{balance}

\newcommand{\parsection}[1]{\vspace{1mm}\noindent\textbf{#1 }}
\usepackage{bbm}
\usepackage{tabularx}
\usepackage{multirow}
\usepackage{booktabs}   
\usepackage{multicol}
\usepackage[dvipsnames]{xcolor}
\usepackage{witharrows}
\usepackage{color, colortbl}
\definecolor{Gray}{gray}{0.90}

\newcommand{\specialcell}[2][c]{%
  \begin{tabular}[#1]{@{}c@{}}#2\end{tabular}}

\NewExpandableDocumentCommand{\gcmidrule}{ O{} D(){} m }{%
    \arrayrulecolor{lightgray}%
    \cmidrule{#3}%
    \arrayrulecolor{black}%
}

\usepackage[acronym]{glossaries}

\newacronym{DNNs}{DNNs}{Deep Neural Networks}
\newacronym{SLAM}{SLAM}{Simultaneous Localization and Mapping}
\newacronym{AR}{AR}{Augmented Reality}
\newacronym{VisLoc}{VisLoc}{Visual Localization}
\newacronym{SfM}{SfM}{Structure-from-Motion}
\newacronym{FPS}{FPS}{Frames Per Seconds}
\newacronym{DoF}{DoF}{Degrees of Freedom}
\newacronym{NeRF}{NeRF}{Neural Radiance Field}
\newacronym{PixSfM}{PixSfM}{Pixel-Perfect-SfM}
\newacronym{NeRFs}{NeRFs}{Neural Radiance Fields}
\newacronym{AUC}{AUC}{Area Under the Curve}

\begin{document}
\title{Long-Term Invariant Local Features via Implicit Cross-Domain Correspondences}

\author{
Zador Pataki,
Mohammad Altillawi,
Menelaos Kanakis,
Rémi Pautrat,\\
Fengyi Shen,
Ziyuan Liu,
Luc Van Gool, and
Marc Pollefeys
\thanks{Z. Pataki, R. Pautrat, and M. Pollefeys are with the Computer Vision and Geometry lab, Department of Computer Science, ETH Zurich}
\thanks{M. Altillawi is with the Computer Vision Center CVC-Barcelona and the Intelligent Robotics Cloud Technology lab of Huawei-Munich}
\thanks{M. Kanakis is with the Computer Vision Lab, Department Electrical Engineering, ETH Zurich}
\thanks{F. Shen is with the Chair of Robotics, Artificial Intelligence and Embedded Systems at TU Munich (TUM) and the Intelligent Robotics Cloud Technology lab of Huawei-Munich}
\thanks{Ziyuan Liu is with the Intelligent Robotics Cloud Technology lab of Huawei-Munich}
\thanks{L. Van Gool is with the Center for Processing Speech and Images, KU Leuven and the Computer Vision Lab, ETH Zurich}

}

\maketitle

\begin{abstract}
Modern learning-based visual feature extraction networks perform well in intra-domain localization, however, their performance significantly declines when image pairs are captured across long-term visual domain variations, such as different seasonal and daytime variations. In this paper, our first contribution is a benchmark to investigate the performance impact of long-term variations on visual localization. We conduct a thorough analysis of the performance of current state-of-the-art feature extraction networks under various domain changes and find a significant performance gap between intra- and cross-domain localization. We investigate different methods to close this gap by improving the supervision of modern feature extractor networks. We propose a novel data-centric method, Implicit Cross-Domain Correspondences (iCDC). iCDC represents the same environment with multiple Neural Radiance Fields, each fitting the scene under individual visual domains. It utilizes the underlying 3D representations to generate accurate correspondences across different long-term visual conditions. Our proposed method enhances cross-domain localization performance, significantly reducing the performance gap. When evaluated on popular long-term localization benchmarks, our trained networks consistently outperform existing methods. This work serves as a substantial stride toward more robust visual localization pipelines for long-term deployments, and opens up research avenues in the development of long-term invariant descriptors.

\end{abstract}


\section{Introduction}
\label{sec:intro}

\begin{figure}[t!]
 \centering
  \includegraphics[width=\linewidth]{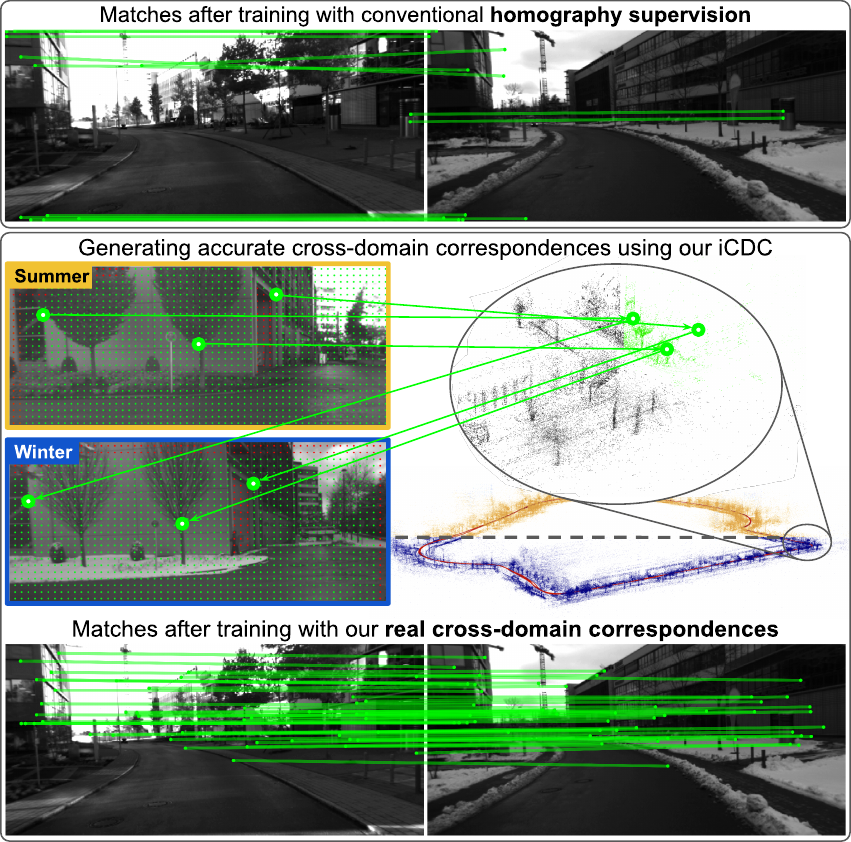}
  
\caption{\textbf{We propose iCDC, a method to supervise feature extractors across different domains.} Traditionally, correspondences are generated by augmenting, and applying homographic transformations to images. iCDC generates accurate correspondences across images captured under different views and in different visual domains.}
  \label{fig:teaser}
\end{figure}

\IEEEPARstart{V}{isual} localization acts as a foundation for many applications, ranging from the navigation of autonomous vehicles to the association of the real and virtual worlds inside, such as placing virtual objects in the user’s field of view. A crucial step in tackling the visual localization problem is data association, where salient points in images are detected and described to establish keypoint correspondences. These correspondences facilitate relative pose estimation of images across varied viewpoints and enable the 3D reconstruction of scenes and the estimation of the 6 \gls{DoF} of the camera pose of a query image relative to a 3D map.

The quality of data association heavily relies on the accuracy and robustness of local feature extraction and description. Initially, sparse features were identified using hand-crafted methods \cite{lowe2004distinctive, bay2006surf, leutenegger2011brisk, rublee2011orb}, which have seen success in various applications, including \gls{SfM}~\cite{colmap} and \gls{SLAM}~\cite{mur2015orbslam}. However, significant advancements in \gls{DNNs}, have enabled the learning of robust, and highly descriptive image features~\cite{superpoint,revaud2019r2d2,dusmanu2019d2,kanakis2023zippypoint,gleize2023silk}, pushing the boundaries of what was previously possible \cite{sattler2018benchmarking, jin2021phototourism, balntas2017hpatches}.

With the ultimate goal being the successful life-long deployment in the real world, robustness to visual changes caused by time of day and season is evaluated using benchmarks such as long-term localization~\cite{sattler2018benchmarking}, or sparse \gls{SLAM} under different weather conditions~\cite{4seasons}. In life-long deployment scenarios, accurate 6 \gls{DoF} localization relies on localization with respect to a prior 3D map. Herein, the key challenge comes from the often differing visual domains of the descriptors in the map and the query. While model-centric methods explored in the state of the art are crucial, we hypothesize that further exploration of methods to better exploit available data (also known as data-centric methods) is necessary to achieve accurate localization performance across long-term variations. 

In this work, we aim to understand and better optimize the robustness of feature extraction networks to long-term visual domain changes. To conduct a thorough evaluation of long-term localization performance, we enhance the 4Seasons dataset~\cite{4seasons} by refining the poses. This dataset contains a large amount of data captured along trajectories across various visual domains and scenes; it lacks consistency in the camera poses across the different trajectories. By refining poses in a cross-trajectory and cross-domain optimization problem, we can accurately compare the relative poses of frames across trajectory and domain pairs of scenes. Using our refined maps, we benchmark current extractors ~\cite{superpoint, revaud2019r2d2, dusmanu2019d2, tang2019neuralKP2D, gleize2023silk}.

Our experiments reveal a significant performance gap between intra-domain and cross-domain localization, regardless of the domain types. To bridge this gap, we explore and categorize data-centric approaches by supervising feature extractors using cross-domain correspondences. Through a comparative analysis, we identify the limitations of the approaches and propose a novel data-centric method, Implicit Cross-Domain Correspondences (iCDC). By representing an environment using multiple, independent, and domain-specific Neural Radiance Fields (NeRFs), iCDC leverages their underlying 3D representations and generates highly accurate cross-domain correspondences. We train state-of-the-art networks \cite{revaud2019r2d2, gleize2023silk} using iCDC, and improve their cross-domain localization performances on our benchmark, and on other long-term localization benchmarks \cite{aachendn, robotcar, cmu}.

In summary, our contributions are the following:
\begin{itemize}
    \item We refine the ground-truth maps of 4Seasons, and establish a new benchmark for long-term localization.
    \item Our benchmarking of existing feature extraction networks reveals a significant performance gap between intra- and cross-domain localization.
    \item We find that supervising extractors using cross-domain correspondences between real viewpoint changes reduces the performance gap.
    \item We propose iCDC to generate accurate cross-domain correspondences. Training extractors using iCDC further improves cross-domain localization.
\end{itemize}

\section{Related Work}
\label{sec:relwork}

\parsection{Hand-crafted feature extractors.} 
Hand-crafted extractors such as SIFT~\cite{lowe2004distinctive}, SURF~\cite{bay2006surf}, BRISK~\cite{leutenegger2011brisk}, BRIEF~\cite{calonder2010brief}, ORB~\cite{rublee2011orb} aim to identify and describe distinctive structures or patterns in images. 
SIFT is well-regarded for its invariance to scale, rotation, and affine transformations, and has been widely used in applications such as object recognition~\cite{lowe2004distinctive} and ~\gls{SfM}~\cite{colmap}. Inspired by SIFT, SURF improved the computation speed of the feature extraction while maintaining robustness to scale and rotation changes, by approximating the Laplacian of Gaussian with a box filter representation. On the other hand, some methods \cite{leutenegger2011brisk,calonder2010brief,rublee2011orb} focus instead on the generation of binary descriptors for use in computationally limited platforms such as robots and \gls{AR}.
While these hand-crafted methods have been successful in a range of applications, they often struggle with perceptual changes caused by variations in illumination, weather, and seasons~\cite{sattler2018benchmarking}.

\parsection{Learned feature extractors.} 
Motivated by the limitations of hand-crafted methods and the success of \gls{DNNs}, learning based feature extractors~\cite{superpoint,revaud2019r2d2,dusmanu2019d2,kanakis2023zippypoint} have improved performance, more noticeably in the aforementioned challenging perceptual changes~\cite{sattler2018benchmarking}.
A number of works explored better architecture designs to enhance performance.
Notably, R2D2~\cite{revaud2019r2d2} and D2-Net~\cite{dusmanu2019d2} propose networks that share all parameters between detection and description.
SuperPoint~\cite{superpoint}, and KP2D~\cite{tang2019neuralKP2D}, jointly learn interest point detection and description using two decoding branches.
Furthermore, other works aimed to improve the optimization of the networks through outlier rejection~\cite{tang2019neuralKP2D}, introducing a novel reliability score to guide the detection process~\cite{revaud2019r2d2}, contrastive learning~\cite{choy2016universal}, using guidance from SfM to obtain more robust keypoints~\cite{tillawi2022pixselect}, or by employing reinforcement learning principles~\cite{tyszkiewicz2020disk}. Recent works \cite{kanakis2023zippypoint,fathy2018hierarchical} investigated the possibility of making these networks more efficient for deployment in computationally limited platforms, while others  \cite{gleize2023silk} focused on improving retrainability; a limitation of DISK~\cite{tyszkiewicz2020disk} which is notoriously difficult to train.

A common component amongst these works, with the exception of DISK, is the use of self-supervised methods, which facilitate the flexible optimization of the networks, enabling learning from diverse and large-scale datasets, such as COCO~\cite{lin2014microsoft}. 
This is accomplished by first applying non-geometric transformations, such as color augmentations, as well as a homographic transformation to every input image, and then optimizing the network to generate identical descriptors for the same regions in the original and transformed images.
While this method has proven very powerful in practice, invariance to perceptual changes caused by weather and seasonal variations is not directly enforced, potentially hindering their performance under challenging perceptual changes caused by the time of day, season, and weather conditions.
In this work, we aim to evaluate the robustness of state-of-the-art extractors in both intra- and cross-domain scenarios, exploring avenues for further improvement.

\parsection{Invariance to tong-term changes.} 
Due to the importance of life-long deployment of computer vision systems, a number of different approaches have been proposed over the years to close the cross-domain performance gap for different applications.
This includes the optimization of networks by directly supervising cross-domain information through contrastive learning~\cite{arandjelovic2016netvlad,chen2017deep,chen2018learning}.
 This has been explored in image retrieval and place recognition networks that output a single descriptor to represent the image, where generating ground-truth labels is made possible through GPS signals available in large datasets \cite{weyand2020google,arandjelovic2016netvlad}.
For a number of applications, however, generating ground-truth labels is expensive. 
This has given rise to domain adaptation~\cite{farahani2021brief,daume2009frustratingly,csurka2017domain} and generalization~\cite{li2018domain,zhou2021domain} methods, allowing the optimization of networks on datasets where labeled data can be more easily acquired, such as simulations, while minimizing the domain gap with real-life data.
This technique has been widely adopted in semantic segmentation~\cite{shen2023loopda,guo2021metacorrection,shen2023diga,wu2021dannet,bruggemann2023refign} and image classification~\cite{tzeng2017adversarial,kang2019contrastive,french2017self} tasks.

More similar to our work, WarpC~\cite{warpc} employs an unsupervised warping consistency to improve correspondence learning. In contrast, other methods \cite{anoosheh2019night,revaud2019r2d2} use style  transfer~\cite{melekhov2020image-stylization} instead, to alter the domain of the images, such as converting a summer image into a winter image.
We, on the other hand, propose to directly enforce cross-domain similarity of descriptors by leveraging a large-scale dataset that captured the same trajectories at different weather conditions~\cite{4seasons}. This is facilitated through our novel method iCDC, a correspondence generation method for real cross-domain images.

\parsection{Long-term datasets.} 
Existing research has shown feature extractor methods to be highly effective for accurate localization within the same domain \cite{sattler2018benchmarking, hloc}. A growing body of literature seeks to address the challenge of long-term localization in situations that require matching descriptors across different domains. Benchmark datasets like the Aachen Day-Night dataset \cite{aachendn} focus on localizing query night-time images with respect to a day-time map, while RobotCar Seasons \cite{robotcar} benchmarks visual localization across an array of query domains against a single reference map capturing a single domain. Notably, its images are of relatively low quality, with night-time images suffering from significant motion blur. On the other hand, CMU-Seasons \cite{cmu} benchmarks localization across many visual domains, albeit primarily in rural areas, with large portions of the images capturing vegetation. Although these benchmarks contribute to the research on cross-domain localization, the range of domain pairs and scenes examined remains limited.

On the other hand, the 4Seasons dataset~\cite{4seasons} encompasses data across multiple trajectories, captured under a variety of weather, seasonal conditions, and in diverse scenes across different environments. However, it lacks reliable relative poses between images captured along different trajectories. To establish it as a new long-term localization benchmark, we jointly optimize the trajectories to provide accurate relative poses across a broad spectrum of visual domains.
\section{Optimizing the 4Seasons Dataset to Benchmark  Cross-Domain Visual Localization}
\label{sec:dataset}

\subsection{Original 4Seasons Dataset Details}
\newcolumntype{C}[1]{>{\centering\arraybackslash}p{#1}}
\begin{table}[ht]
\setlength\tabcolsep{9pt}
  \center
  
    \caption{\textbf{Dataset Overview.} Trajectory labels are used in this work to refer to the corresponding trajectories. The domain is described for each trajectory. For each scene, the average number of frames with RTK-GNSS measurements per trajectory is provided, as well as the average of the total number of available frames.}
      \footnotesize
 \begin{tabular}{@{}C{1.4cm}C{1.7cm}C{1.7cm}C{1.7cm}@{}}\toprule
\centering \specialcell{Trajectory \\ Labels} & Season & \specialcell{Weather  \\ Condition} & Time of Day\\ 
\midrule
\multicolumn{4}{l}{\textbf{Business Campus} (3255 RTK-GNSS poses, 11765 available frames)}\\
BC1  &  	Fall & Sunny & Morning  \\
BC2         &  Winter & Cloudy/snowy & Afternoon \\
BC3        &  	Winter & Sunny & Afternoon  \\
\midrule
\multicolumn{4}{l}{\textbf{Office Loop} (2775 RTK-GNSS Poses, 14438 available frames)}\\
OL1  &  Spring & Sunny & Afternoon  \\
OL2  &  Spring & Sunny & Afternoon \\
OL3  &  Spring & Sunny & Morning  \\
OL4  &  Summer & Sunny & Morning  \\
OL5  &  Winter & Cloudy/snowy & Afternoon \\
OL6  &  Winter & Sunny & Afternoon  \\
\midrule
\multicolumn{4}{l}{\textbf{Neighborhood} (2790 RTK-GNSS poses, 10297 available frames)}\\
N1  &  	Spring & Cloudy & Afternoon  \\
N2  &  	Fall & Cloudy & Afternoon \\
N3  &  Fall & Rainy & Afternoon  \\
N4  &  Winter & Cloudy & Morning  \\
N5  &  Winter & Sunny & Afternoon \\
N6  &  Spring & Cloudy & Evening  \\
N7  &  Spring & Cloudy & Evening  \\
\bottomrule
\end{tabular}
\label{tab:4seasons dataset details}
\end{table}

Building on existing datasets \cite{aachendn, robotcar, cmu}, Sattler et al. \cite{sattler2018benchmarking} established the benchmark of long-term visual localization. Compared to the existing datasets in this benchmark, the autonomous driving 4Seasons dataset~\cite{4seasons} provides data with accurate ground truth poses along multiple trajectories under a more varied set of weather, seasonal, and day-time conditions across a larger number of distinct environments. For our cross-domain localization benchmark, the first criterion for the dataset was having a wide range of categorical variations in visual conditions, which the 4Seasons dataset satisfied. However, for a comprehensive analysis, the dataset must fulfill an additional criterion: ground truth relative poses must be accurate across individual trajectories, i.e. across visual domains. Together, a dataset satisfying these two criteria would provide the flexibility of evaluating the localization performances of networks across any two visual conditions available in a scene.

In the 4Seasons dataset, Real-time Kinematic and Global Navigation Satellite System (RTK-GNSS) measurements were utilized for a subset of frames to enhance the accuracy of the ground truth poses. For further details, please refer to the 4Seasons paper \cite{4seasons}. Additionally, these measurements provide geolocations for each frame, enabling the representation of individual trajectories within a unified coordinate system. However, the pose optimization within the dataset was conducted solely along individual trajectories, and as a result, the available maps of the 4Seasons dataset do not satisfy our second criterion. The existing cross-trajectory poses in the dataset are not reliable for benchmarking localization performance. Moreover, the unreliable relative poses result in misaligned local 3D representations of the maps, making them unsuitable for our data-centric supervision approach. This underscores the necessity for improved maps.

\subsection{Refining the 4Seasons Dataset}
\begin{figure*}[t!]
 \centering
  \includegraphics[width=\linewidth]{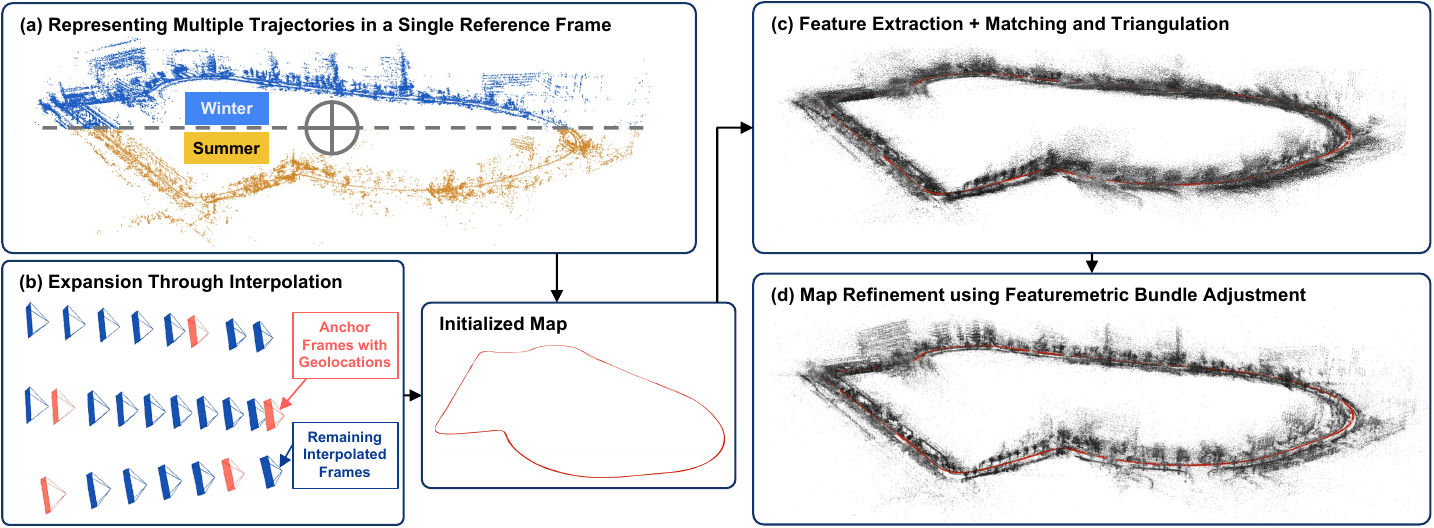}
  
\caption{\textbf{Joint Trajectory Map Refinement.} (a) Frames with registered geolocations along multiple trajectories of the 4Seasons dataset are placed, as anchors, into a global map. (b) The remaining frames have poses available only in a local reference frame optimized through stereo visual-inertial odometry. They are integrated into the map by interpolating them around the anchor frames. (c) SuperPoint \cite{superpoint} features are extracted and matched using SuperGlue \cite{superglue}. Using the poses in the initialized map, the matched keypoints are triangulated into 3D. (d) Pixel-Perfect-SfM \cite{pixsfm} is leveraged to optimize the 3D points and poses in the map through featuremetric bundle adjsutment. }
  \label{fig:maps}
\end{figure*}

We follow the HLoc \cite{hloc} pipeline, which integrates and enhances COLMAP \cite{colmap}, to build SfM maps that satisfy our criteria for a subset of the scenes available in the 4Seasons dataset. In Tab. \ref{tab:4seasons dataset details}, we present an overview of the data from these scenes, including trajectory labels referred to later in this paper. To accelerate the map optimization process, we use the available poses in the 4Seasons dataset as initialization. Our refinement pipeline is summarized in Fig. \ref{fig:maps}.

\parsection{Map initialization.} We initialize our maps using the provided poses of all frames of a 4Seasons scene, across multiple trajectories in a unified coordinate system. All frames with registered geolocations, i.e. with RTK-GNSS positions, are placed in the unified coordinate system and serve as anchors for map initialization. For the remaining frames, poses are available only in a local coordinate system, optimized through stereo visual-inertial odometry. Despite suffering from drift, we use the locally accurate relative poses to initialize the global map by placing them around the anchor frames. 

\parsection{Image matching.} In the first step of the HLoc pipeline, covisible database images are identified. We generate these covisible image pairs using the ground truth relative poses available in our initialized maps. Per frame, we find the nearest $n$ frames inter- and intra-trajectory. 

\begin{figure*}[t!]
 \centering
  \includegraphics[width=0.83\linewidth]{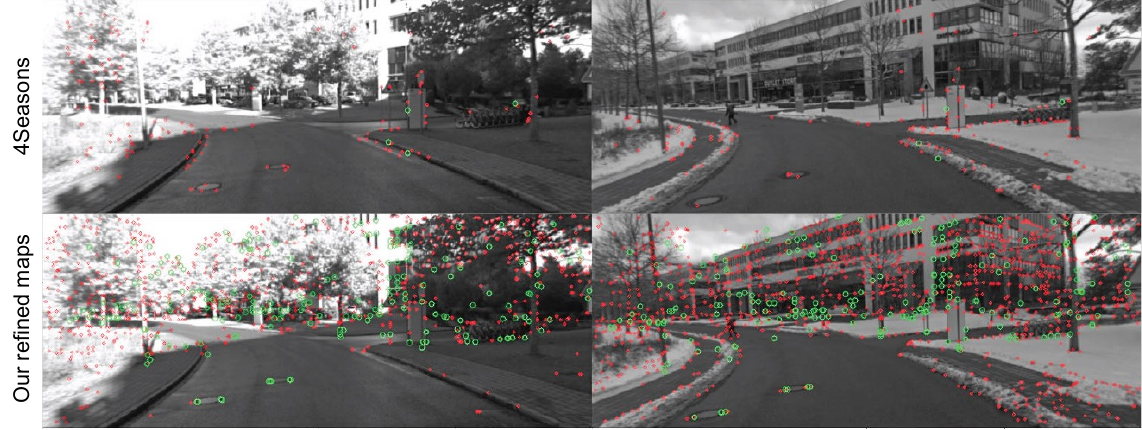}
  
\caption{\textbf{Keypoints and Matches in 4Seasons Dataset and Our Refined Maps.} The figure showcases keypoints from the 4Seasons dataset and from our refined maps. Matches across frames are highlighted in green. For the 4Seasons dataset, keypoints are matched using corresponding depth values and frame poses, while matches from our refined maps are directly extracted from the maps.}
  \label{fig:map_matches}
\end{figure*}

\parsection{SfM map triangulation.} In the next stage, COLMAP triangulates shared keypoints across images into 3D. Although the provided 4Seasons poses can be used to initialize the maps, the provided keypoints can not be relied on for global map optimization. Specifically, there is only a limited number of available keypoints and the dataset does not provide keypoint matches across image pairs. Per-keypoint depth values, when combined with relative poses, could theoretically facilitate geometric keypoint matching. However, the unreliability of relative poses across trajectories makes these geometric keypoint matches unreliable as well. Instead, we detect keypoints using SuperPoint \cite{superpoint}, match them across identified image pairs using SuperGlue \cite{superglue}, and use them for triangulation. Fig. \ref{fig:map_matches} compares the sparse set of cross-domain 4Seasons keypoints and geometrical matches with the abundance of SuperPoint + SuperGlue matches in our maps. SuperPoint + SuperGlue generates accurate matches across domains \cite{aachendn}, and in a SfM pipeline, generates highly accurate maps.

\parsection{Global map optimization.} In the final stage of the HLoc pipeline, maps are globally optimized through a process of bundle  adjustment. Conventionally, COLMAP would be used to refine the poses and 3D points in the map by globally minimizing reprojection errors. Taking advantage of the latest advances in SfM, we utilize Pixel-Perfect-SfM (PixSfM) \cite{pixsfm}. In contrast to COLMAP, the global refinement is coined as ``featuremetric", because it optimizes the consistency of deep features across frames, and it additionally optimizes keypoints. Instead of relying on reprojection errors, 3D points, keypoints, and poses are optimized by minimizing a featuremetric cost.  

\subsection{Refined Map Quality Analysis}
Due to the absence of ``true" ground truth poses, we can not quantify the true accuracy of our maps, however, we demonstrate the improved quality of our maps over the ground truth poses available in the 4Seasons dataset. We provide both a qualitative and a quantitative analysis. Although we generate relative poses across all frame pairs in a scene, for a comprehensive analysis, we limit this study to the frames with registered geolocation positions; in the 4Seasons dataset, the other frames exist in their own local coordinate system. 

\parsection{Qualitative map quality analysis.}
We reproject keypoints from a query image onto a reference image using the estimated depths and the relative poses from the corresponding 3D map. The quality of the maps can be visually investigated by studying the reprojection of keypoints of a static object, such as a building. Errors in the map reveal themselves when the reprojected points on the query image do not match the same points in the reference image. These errors can be rooted in errors in both depth and pose, however, in contrast to depth errors, pose errors yield consistent reprojection errors across all keypoints in the image pair. 

\begin{figure}[t!]
 \centering
  \includegraphics[width=\linewidth]{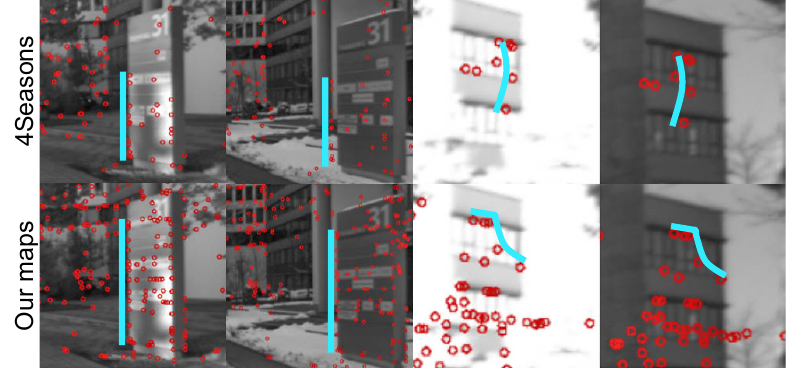}
  
\caption{\textbf{Keypoint Reprojection for Pose Quality Analysis.} Each row presents image pairs captured during different seasons.  Keypoints from each left-side image are reprojected onto the right-side image using ground truth poses and keypoint depths. Top and bottom rows are reprojected keypoints from the 4Seasons maps and our maps respectively. Lines have been drawn to highlight reference and corresponding reprojected keypoints. The projected keypoints along the highlighted lines reveal consistent reprojection errors across the 4Seasons frame pairs. These consistent errors indicate inaccurate relative poses. The same can not be observed in the case of our refined maps.}
  \label{fig:reprojection}
\end{figure}

In Fig. \ref{fig:reprojection}, we showcase cross-domain pairs of reference and query images that demonstrate the reprojection of keypoints of static objects and reveal relative pose errors in the 4Seasons dataset. These can be clearly seen along edges of the static objects (we highlight some with cyan lines). In contrast to reprojecting using keypoints and poses from the original 4Seasons dataset, reprojection using the refined poses yields more accurate correspondences.

\parsection{Quantitative map quality analysis.}
In the next section, we additionally benchmark the cross-domain localization performances of state-of-the-art feature extraction networks using our refined poses as ground truth, and properly introduce the metrics. Here, we present a summary of the results in Tab. \ref{tab:4seasonsvsrefined}, in addition to the same localization results, evaluated against the 4Seasons poses treated as ground truth. We report the median localization errors and, following \cite{superglue, sun2021loftr}, \gls{AUC} of pose errors at error thresholds: 5$^\circ$, 10$^\circ$, and 20$^\circ$, averaged across all scenes and trajectory pairs.

All feature extractors achieve higher performances, i.e. lower median errors and higher \gls{AUC} under all thresholds, when benchmarked against our refined poses. Due to their inherent imperfections, extractor performances are not a perfect indicator of improved map accuracies, however, the consistent disparity in localization errors emphasizes the enhancements achieved during our refinement. 
While the initial goal was to improve cross-domain poses, our experiment also reveals enhanced ground-truth poses along individual trajectories.

The qualitative and quantitative analysis combined highlights that the refined poses can be dependably used for evaluating cross-domain localization accuracies. This new dataset is used from here on for the remainder of the paper. With the capability to evaluate localization performances across all visual domains available in each scene, our presented benchmark offers an opportunity for the most comprehensive study of cross-domain localization to date.

\begin{table*}[t!]
\setlength\tabcolsep{1.9pt}
  \center
  
    \caption{\textbf{Relative Pose Errors Evaluated Against 4Seasons Poses vs. Our Refined Poses.} Pose errors are presented both as median errors and AUC accuracies. Consistent higher performances when evaluated against our refined poses indicates higher ground truth pose quality.}
      \scriptsize
 \begin{tabular}{@{}crccccccccccc@{}}\toprule
\label{tab:4seasonsvsrefined}
&& \multicolumn{5}{c}{\textbf{Same Domain}} & &\multicolumn{5}{c}{\textbf{Cross Domain}} \\
\cmidrule{3-7} \cmidrule{9-13}
& & \specialcell{SuperPoint \cite{superpoint}} & \specialcell{R2D2 \cite{revaud2019r2d2}} & \specialcell{D2-Net \cite{dusmanu2019d2}} & \specialcell{KP2D \cite{tang2019neuralKP2D}} &  \specialcell{SiLK \cite{gleize2023silk}} && \specialcell{SuperPoint \cite{superpoint}} & \specialcell{R2D2 \cite{revaud2019r2d2}} & \specialcell{D2-Net \cite{dusmanu2019d2}} & \specialcell{KP2D \cite{tang2019neuralKP2D}}  &  \specialcell{SiLK \cite{gleize2023silk}}\\
 \midrule
\multirow{2}{*}{\specialcell{\textbf{Median} \\\textbf{Error ($^\circ$) $\downarrow$}}}&\specialcell{4seasons} & 1.05 & 0.90 & 1.57 & 1.00 & 0.97 && 9.94 & 13.07 & 10.13 & 12.44 & 17.98 \\
&\specialcell{Refined} & \textbf{0.96} &  \textbf{0.81} & \textbf{1.49} & \textbf{0.91} & \textbf{0.88} && \textbf{9.78} & \textbf{12.65} & \textbf{9.89} & \textbf{12.23} & \textbf{17.84} \\
 \midrule
\multirow{2}{*}{\specialcell{\textbf{AUC} \\ (\%) $\uparrow$}}&\specialcell{4seasons} & 71.6/84.4/91.5 & 76.2/87.4/93.4 & 60.7/77.7/87.5 & 73.9/85.7/91.9 & 74.5/86.1/92.3 && 27.4/41.8/54.5 &28.4/41.8/52.9 & 21.8/37.4/52.1 & 26.5/40.1/51.9 & 23.1/35.6/47.2 \\
&\specialcell{Refined} & \textbf{73.7}/\textbf{85.7}/\textbf{92.3} & \textbf{78.5}/\textbf{88.7}/\textbf{94.1} & \textbf{62.4}/\textbf{78.7}/\textbf{88.1} & \textbf{76.0}/\textbf{86.9}/\textbf{92.7} & \textbf{76.3}/\textbf{87.0}/\textbf{92.7} && \textbf{29.1}/\textbf{43.0}/\textbf{55.2} & \textbf{30.6}/\textbf{43.5}/\textbf{54.1} & \textbf{22.8}/\textbf{38.1}/\textbf{52.5} & \textbf{28.2}/\textbf{41.3}/\textbf{52.6} & \textbf{24.5}/\textbf{36.4}/\textbf{47.6}\\

\bottomrule
\end{tabular}
\label{table:depth_joint_perf}
\end{table*}

\section{Benchmarking Localization Robustness Across Visual Domains}
\label{sec:benchmarking}
Utilizing our refined maps, we benchmark state-of-the-art feature extraction methods in the task of long-term localization. 
We evaluate localization performance using both relative and absolute pose estimation. 

\subsection{Benchmarking Relative Pose Estimation} 
To benchmark feature extractors, we propose to match keypoints between image pairs, estimate the relative pose between the two frames, and evaluate the quality of the predicted pose.
The estimated translation can only be determined up to a scale factor and is therefore reported as a normalized direction vector. Rotation and translation pose errors are the relative angles between estimated and ground truth rotation matrices and translation vectors. Following previous works \cite{superglue, sun2021loftr}, we report the pose errors, in degrees, as the maximum of translation and rotation angular errors.

\parsection{Keyframing.} 
For each scene and each trajectory pair, we select a set of reference frames and corresponding query frames. We select frames at increments of 10 meters along the reference trajectory. For each selected reference frame, we select corresponding query frames from the set of frames that are within 8 meters and within 45$^\circ$ of the reference frame. We select these query frames at increments of 2 meters.

\newcommand{\mc}[2]{\multicolumn{#1}{c}{#2}}
\definecolor{Gray}{gray}{0.85}
\newcolumntype{a}{>{\columncolor{Gray}}c}

\begin{table*}[ht]
    \setlength\tabcolsep{5.1pt}
  \center
    \caption{\textbf{Median Relative Pose Errors on the 4Seasons Dataset \cite{4seasons}.} For each scene, aggregated median errors are presented, evaluated against our ground truth poses. In the case of cross-domain, the error under reference trajectory is the mean of all median pose errors between the reference and all query trajectories.} 

      \scriptsize
 \begin{tabular}{@{}rrrrcrrrrrrcrrrrrrrca@{}}\toprule
 
\textbf{Scenes}& \multicolumn{3}{c}{Business Campus} & \phantom{}& \multicolumn{6}{c}{Office Loop} &
\phantom{} & \multicolumn{7}{c}{Neighborhood}\\
\cmidrule{2-4} \cmidrule{6-11} \cmidrule{13-19}
\textbf{Traj.}& BC1 & BC2 & BC3  && OL1 & OL2 & OL3 & OL4& OL5& OL6  && N1& N2 & N3& N4 & N5 & N6& N7 && ALL\\ \midrule
\multicolumn{2}{l}{\textbf{Intra-domain localization}}\\
SuperPoint~\cite{superpoint}  & 1.28 & 1.07 & 1.26  &&  1.06 & 1.02 & 0.95 & 0.91 & 0.90 & 0.94  && 0.85 & 0.78 & 0.89 & 0.90 & 0.79 & 0.85 & 0.88  &&  0.96\\
R2D2~\cite{revaud2019r2d2}         & \textbf{1.03} & 0.93 & \textbf{0.99}  &&  \textbf{0.84} & \textbf{0.81} & \textbf{0.82} & \textbf{0.80} & \textbf{0.76} & \textbf{0.76} &&  \textbf{0.72} & \textbf{0.71} & \textbf{0.78} & \textbf{0.78} & \textbf{0.71} & \textbf{0.75} & \textbf{0.76}  &&  \textbf{0.81}\\
D2-Net~\cite{dusmanu2019d2}       &  2.13 & 1.56 & 1.70 && 1.65 & 1.65 & 1.55 & 1.39 & 1.38 & 1.45 && 1.35 & 1.28 & 1.41 & 1.52 & 1.21 & 1.26 & 1.34 && 1.49\\
KP2D~\cite{tang2019neuralKP2D}       &  1.08 & \textbf{0.91} & 1.05  &&  0.99 & 0.96 & 0.98 & 0.92 & 0.84 & 0.93  && 0.85 & 0.79 & 0.87 & 0.90 & 0.83 & 0.83 & 0.88  &&  0.91 \\
SiLK \cite{gleize2023silk}        &  1.11 & 0.88 & 0.99  &&  0.96 & 0.96 & 0.93 & 0.83 & 0.83 & 0.86  && 0.82 & 0.76 & 0.86 & 0.90 & 0.77 & 0.83 & 0.83  &&  0.88 \\
\multicolumn{2}{l}{\textbf{Cross-domain localization}}\\
SuperPoint~\cite{superpoint}  &  \textbf{13.94} & \textbf{9.51} & \textbf{9.70}  &&  14.02 & 15.08 & 10.37 & 35.37 & 14.86 & 13.65  && 2.68 & \textbf{3.02} & \textbf{3.02} & 2.48 & \textbf{4.13} & 2.22 & 2.38  &&  \textbf{9.78}\\
R2D2~\cite{revaud2019r2d2}        &  20.02 & 12.44 & 13.29  &&  17.95 & 19.33 & 11.40 & 50.52 & 18.06 & 18.79  && \textbf{2.63} & 3.19 & 3.56 & \textbf{2.27} & 4.64 & \textbf{2.07} & \textbf{2.22}  && 12.65 \\
D2-Net~\cite{dusmanu2019d2}       &  14.99 & 10.54 & 10.23 && \textbf{13.66} & \textbf{13.63} & \textbf{8.86} & \textbf{32.66} & \textbf{13.02} & \textbf{12.26} && 3.99 & 4.03 & 4.32 & 4.13 & 4.93 & 3.35 & 3.60 && 9.89 \\
KP2D~\cite{tang2019neuralKP2D}         &  19.90 & 13.34 & 14.37  &&  16.58 & 16.11 & 11.36 & 46.94 & 16.70 & 16.52  && 3.05 & 3.94 & 3.62 & 2.57 & 5.63 & 2.40 & 2.65  &&  12.23\\
SiLK \cite{gleize2023silk}        &  19.76 & 12.03 & 14.09  &&  33.15 & 33.36 & 19.46 & 60.70 & 24.80 & 26.54  && 3.68 & 5.08 & 5.39 & 3.15 & 8.78 & 2.88 & 3.53  &&  17.84 \\
\bottomrule
\end{tabular}
\label{tab:relative pose errors}
\end{table*}

\begin{figure*}[t!]
 \centering
  \includegraphics[width=\linewidth]{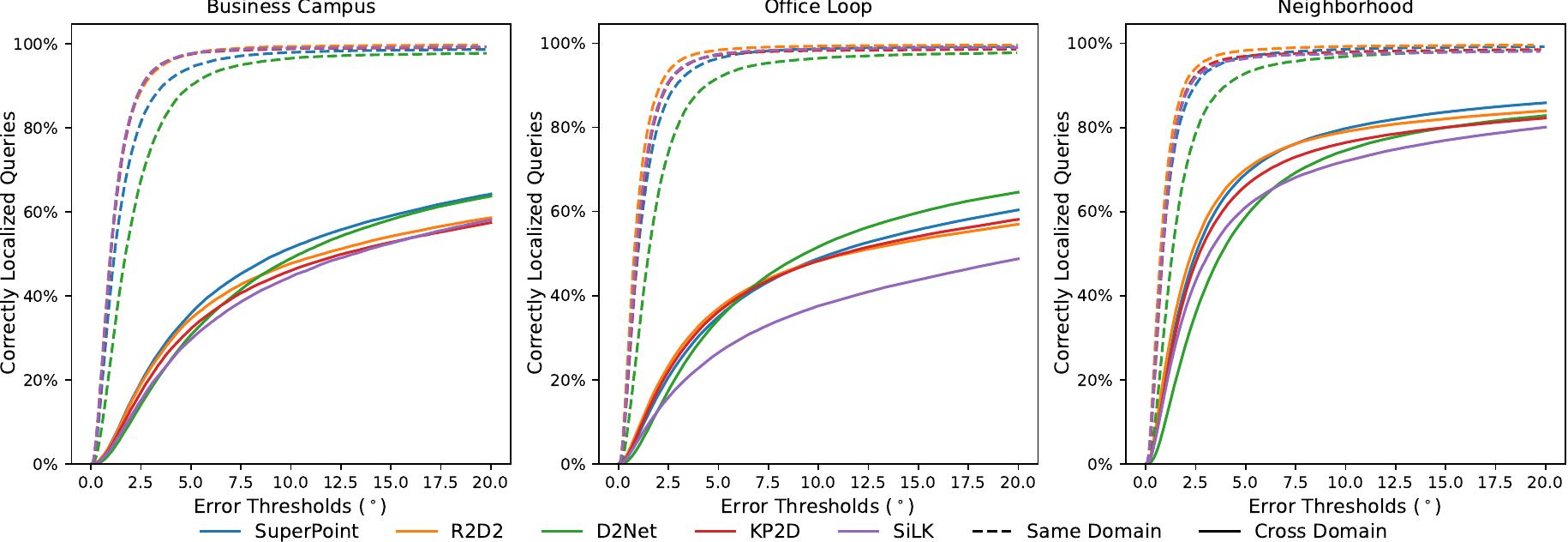}
\caption{\textbf{Cumulative Distribution of Relative Pose Errors on the 4Seasons Dataset \cite{4seasons}.} For each scene, we present the AUC curves separately evaluated against our ground truth poses. Cross-domain plots are marked with filled lines while intra-domain plots are marked with dashed lines. The plots reveal a large performance gap between intra- and cross-domain localization.}
\label{fig:relative AUC}
\end{figure*}

\parsection{Relative pose errors.} 
In some cases, the sampled keyframe pairs have large relative poses. This often results in significant occlusion, particularly across different trajectories, rendering accurate relative localization impossible. To account for such outliers, we calculate median pose errors per trajectory pair. In Tab. \ref{tab:relative pose errors}, we report metrics for both intra-domain and cross-domain cases. In intra-domain localization, relative poses are estimated across image pairs captured along the same trajectory, while in cross-domain localization, they are estimated across different trajectories. In the cross-domain case, the reported errors per reference trajectory are the mean of all median pose errors between the reference and all query trajectories. Furthermore, we plot in  Fig. \ref{fig:relative AUC} the percentage of correctly localized images under different thresholds across all trajectory pairs. While we do not get insight into localization performances along individual trajectories, these plots provide a better understanding of the distribution of localization errors.

In Tab. \ref{tab:relative pose errors}, all feature extraction networks exhibit significantly lower localization errors in intra-domain localization. Furthermore, this performance gap is clearly illustrated in Fig. \ref{fig:relative AUC}. In intra-domain localization, R2D2~\cite{revaud2019r2d2} consistently outperforms other methods, while D2-Net \cite{dusmanu2019d2} struggles. In contrast, D2-Net demonstrates robust results in cross-domain localization for scenes with larger variations, outperforming R2D2, KP2D \cite{tang2019neuralKP2D} and SiLK \cite{gleize2023silk}. Particularly SiLK suffers in cross-domain localization, consistently achieving the worst results across the board, and dramatically lagging behind in Office Loop. The results presented in the table suggest that trajectories BC1 and OL4, both characterized by sunny conditions, pose particularly challenging localization tasks within their respective scenes, likely due to the presence of shadows. These shadows present challenges in cross-trajectory localization by introducing edges in one domain that are absent in the other. The challenges in cross-domain localization also escalate under snowy conditions (as seen in BC3 and OL5), given that snow introduces markedly different textures and leafless trees present new geometries.

\subsection{Benchmarking Absolute Pose Estimation}
We further investigate the cross-domain localization performances of the extractors by calculating the absolute 6 \gls{DoF} pose of query images with respect to a reference map, utilizing the HLoc localization approach \cite{hloc}.  Building on the poses available in our maps, the reference maps are reconstructed using local features extracted by the feature extractor being evaluated.
For a given query image, image retrieval is performed to obtain the $n$ most similar images from the database of the reference images. Rather than searching for matches throughout the 3D model, this process effectively narrows down the search space.

In Tab. \ref{table:hloc benchmaring}, following \cite{sattler2018benchmarking}, we report the percentage of localized images under three thresholds: (0.25m, 2$^\circ$), (0.5m, 5$^\circ$) and (5m, 10$^\circ$). We report the results using two methods of image retrieval. The first method utilizes the ground-truth poses of the images to retrieve the closest reference images to the query. The second method relies on NetVLAD \cite{arandjelovic2016netvlad}.
We compute the localization accuracy per reference-query trajectory pair. For each scene, we present cross-domain localization performances, averaging accuracies over all trajectory pairs.

R2D2 and SuperPoint achieve the highest correctly localized images under the finer thresholds (0.25m/2$^\circ$, 0.5m/5$^\circ$) while D2Net reports the best localization result under the coarse threshold (5m/10$^\circ$). R2D2~\cite{revaud2019r2d2} performs better than SuperPoint~\cite{superpoint} in scenes with smaller domain variations. In contrast, SuperPoint~\cite{superpoint} demonstrates greater robustness to larger variations in viewing conditions that are encountered in the Business Campus and Office Loop scenes.

The results that are obtained using NetVLAD for image retrieval mirror those obtained by using the ground truth poses except that the values are lower. This outcome is expected as using ground-truth pose for image retrieval, facilitates the retrieval of the query images closest to the reference image. Utilizing ground-truth poses in this manner helps in isolating the localization errors from image retrieval errors, with the former stemming from the quality of the reconstructed map, as well as the description and matching of local features. Similarly to the case of relative pose estimation, SiLK struggles in cross-domain localization. KP2D was omitted from this experiment, as it was not implemented in HLoc. 

\begin{table*}[ht]
\setlength\tabcolsep{11.8pt}
  \center
    \caption{\textbf{Cross-domain Absolute Pose Accuracies on the 4Seasons Dataset \cite{4seasons}.} For each scene the percentages of accurately localized query images under thresholds (0.25m, 2$^\circ$), (0.5m, 5$^\circ$), and (5m, 10$^\circ$) are reported, evaluated against our ground truth psoes. Results are presented for the cases when ground truth poses and NetVLAD~\cite{arandjelovic2016netvlad} are used for image retrieval.}
      \scriptsize
 \begin{tabular}{@{}rccccccc@{}}\toprule
 \textbf{Retriaval}& \multicolumn{3}{c}{Ground Truth} & \phantom{}& \multicolumn{3}{c}{NetVLAD} \\
 \cmidrule{2-4} \cmidrule{6-8} 
\textbf{Scenes}& Business Campus & Office Loop & Neighborhood && Business Campus & Office Loop & Neighborhood\\
 \midrule
SuperPoint \cite{superpoint} &  \textbf{66.6} / 82.6 / 99.7         & \textbf{47.2} / \textbf{60.7} / 89.9   & 71.5 / 82.0 / 97.6                     && \textbf{64.5} / 80.3 / 97.3           &  \textbf{44.4} / \textbf{56.1} / 78.6  &  70.5 / 80.6 / 95.3 \\
R2D2 \cite{revaud2019r2d2} &  62.1 / 80.7 / 99.8                    &  43.2 / 58.06 / 88.3                   & \textbf{72.2} / \textbf{83.4} / 98.0   && 60.2 / 78.6 / 98.2                    & 41.3 / 54.7 / 79.9  & \textbf{71.4} / \textbf{81.9} / 96.2  \\
D2-Net \cite{dusmanu2019d2} &  61.8 / \textbf{83.1} / \textbf{99.9}  &  42.9 / 59.1 / \textbf{92.3}           & 65.8 / 80.1 / \textbf{98.5}                     && 59.7 / \textbf{80.9} / \textbf{98.6}  &  40.6 / 55.3 / \textbf{83.1}  &  64.8 / 80.0 / \textbf{96.5} \\
SiLK \cite{gleize2023silk}        &  44.9 / 61.2 / 85.4         & 36.0 / 46.6 / 68.0   & 65.5 / 77.7 / 93.7                    && 42.2 / 57.4 / 80.2           &  33.0 / 42.3 / 59.7 &  64.3 / 76.0 / 90.7 \\
\bottomrule
\end{tabular}
\label{table:hloc benchmaring}
\end{table*}

In all experiments, a noticeable performance gap persists between intra-domain and cross-domain localization across all feature extractor methods, regardless of the domain type. While these extractors exhibit outstanding localization performance in intra-domain tasks, the persistent gap underlines that there is still significant room for improvement to achieve true long-term robustness in descriptors. Current state-of-the-art research has made remarkable strides in refining extractors using model-centric approaches such as improving models and optimization schemes. However, in this paper, we argue that these approaches are not sufficient for bridging this gap, as they do not incorporate real cross-domain variations across correspondences. Instead, we advocate for a shift towards data-centric approaches. In the forthcoming sections, we motivate the need for real cross-domain correspondences for supervising feature extractors to enforce cross-domain invariance. Furthermore, we present iCDC in detail and demonstrate a significant reduction in the performance gap when using its correspondences for supervision.

\section{Network Supervision for Long-Term Invariant Descriptors}
\label{sec:supervision}

We can reduce the previously observed performance gap by supervising feature extractors using real cross-domain correspondences. However, manually annotating correspondences is infeasible, as it requires per-pixel annotations for each image pair. As a result, state-of-the-art methods \cite{revaud2019r2d2, superpoint,tang2019neuralKP2D, gleize2023silk} utilize homographic transformations for flexibly generating correspondences between the reference and the transformed image. However, homographies cannot encapsulate real-world geometric variations, and image augmentations must be relied on to generalize across lighting variations.

\subsection{Correspondence Generation Overview}
Structure-based methods provide a reliable way to generate correspondences across different views \cite{tillawi2022pixselect, larsson2019cross}. Relying solely on the sparse SfM maps, however, only results in a sparse set of correspondences. The initial sparse reconstruction using SfM can be further processed using Multi-view stereo to obtain a denser reconstruction. However, the resulting depth is often noisy and yields suboptimal correspondences.

Dense correspondence estimation is a field where networks estimate dense matches between image pairs, and could be used to generate dense cross-domain correspondences for supervision \cite{revaud2019r2d2}. However, estimating dense correspondences across long-term viewing conditions is particularly challenging due to significant variations in lighting, texture, and geometry.

\begin{figure*}[t!]
    \centering
    \begin{tabular}{@{}cc@{}}
    \includegraphics[width=.3\linewidth]{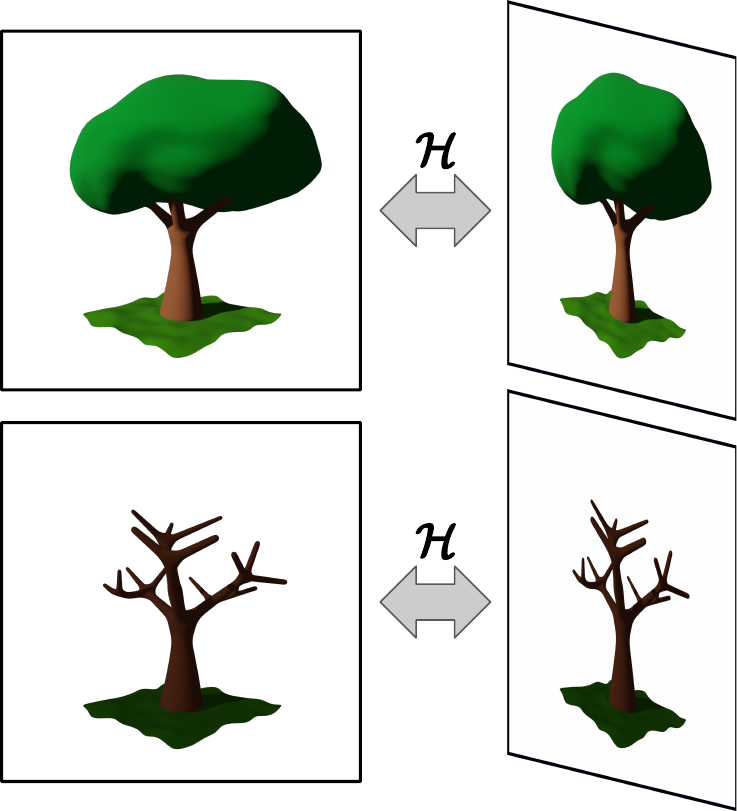} & \includegraphics[width=.4\linewidth]{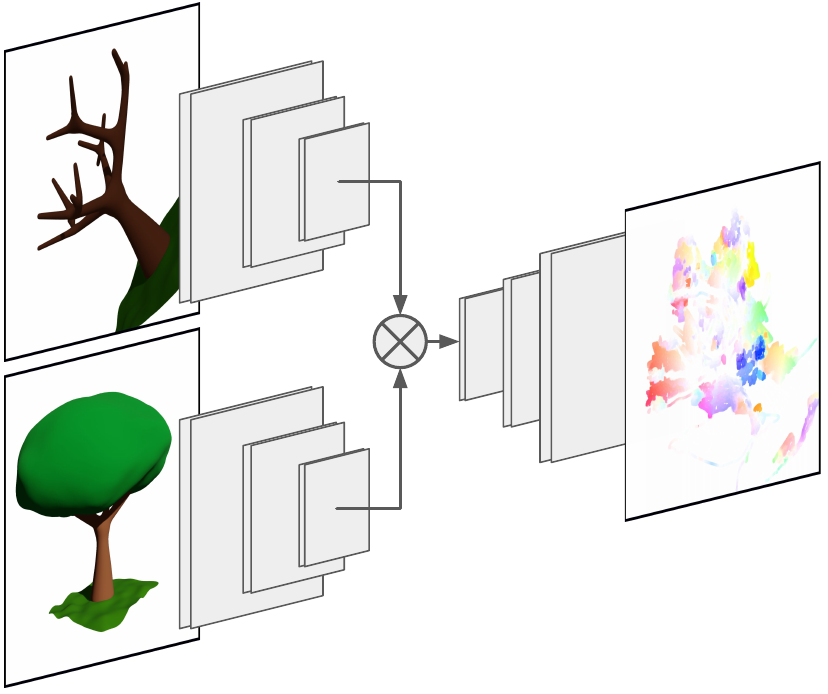} \vspace{-0.1in} \\
    \small (a) Homography & \small (b) Dense Correspondence Network\\
    \includegraphics[width=0.41\linewidth]{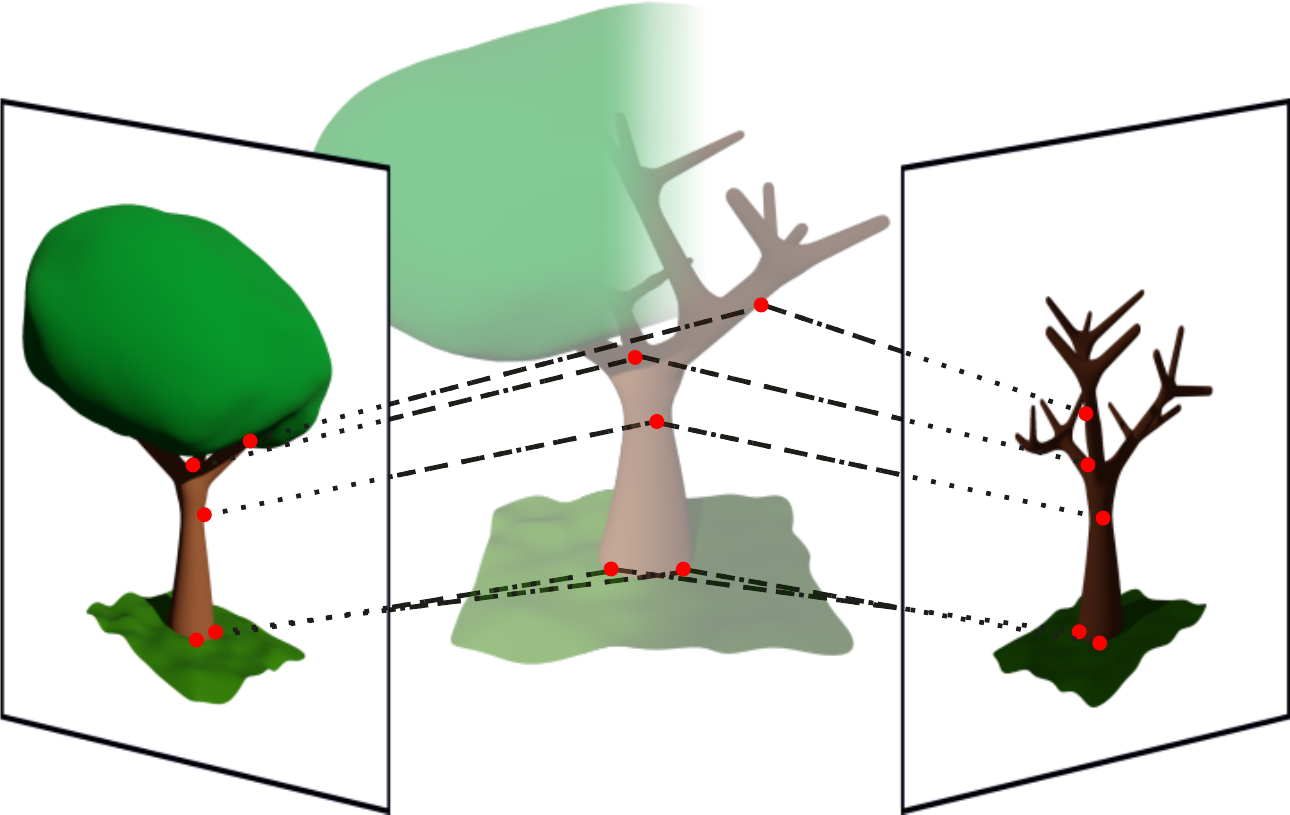} & \includegraphics[width=.41\linewidth]{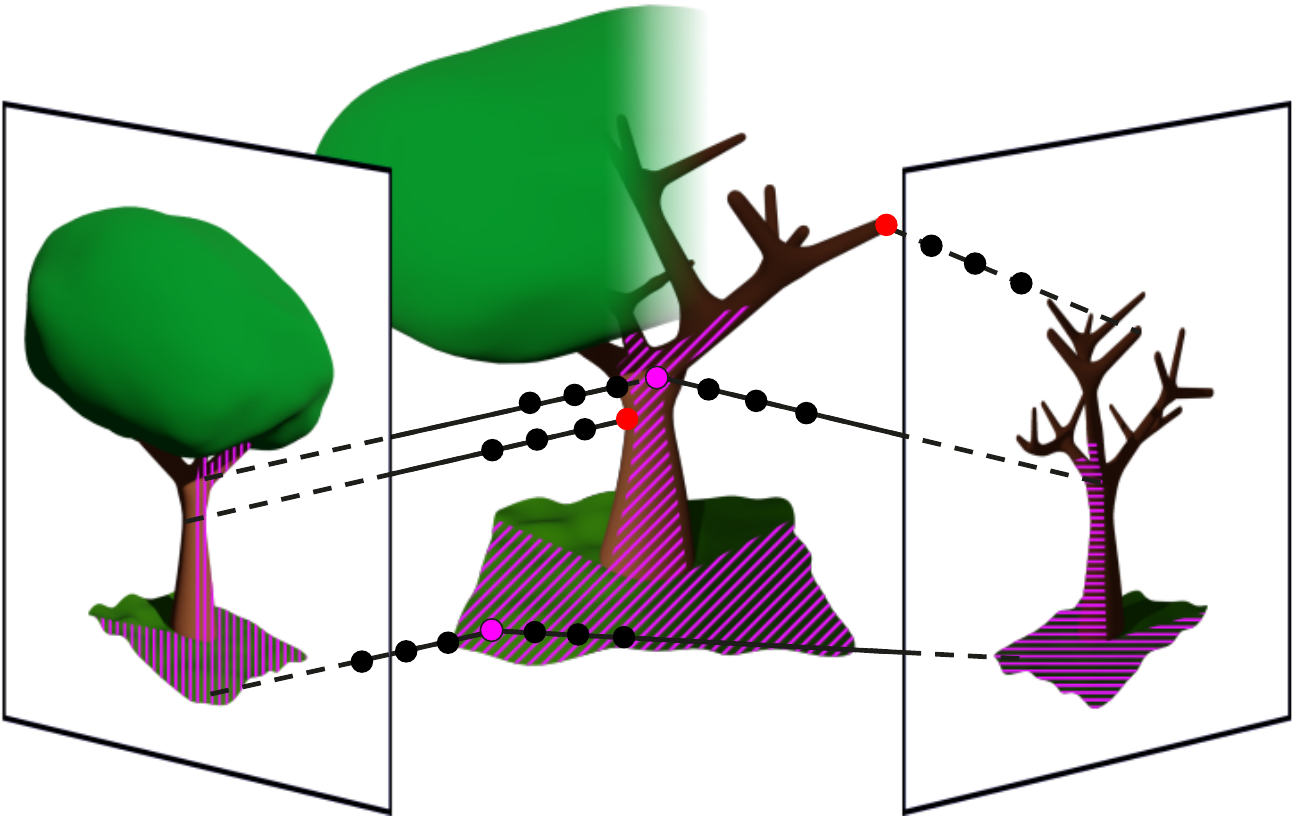}\vspace{-0.1in} \\
    \small (c) Sparse Map & \small (d) iCDC (ours)\\
 
  \end{tabular}

    \caption{\textbf{Overview of Explored Correspondence Methods.} Homographic transformations (a) are used to train most extractor methods out of convenience. Dense correspondence estimation networks (b) can estimate correspondences between real-world images across arbitrary viewpoint changes. However, these methods can be inaccurate, particularly across visual domains. Sparse \gls{SfM} map matches (c) can serve as correspondences, providing highly accurate sparse correspondences across domains and views. In our method iCDC (d), we train domain-specific NeRFs using cross-trajectory \gls{SfM} maps, and generate correspondences by comparing the implicit 3D geometries. Highlighted regions are where correspondences are extracted.}
    \label{fig:joined}
\end{figure*}

In Fig. \ref{fig:joined}, we present an overview of all correspondence generation methods we explore. In addition to using homographies, we use a dense correspondence network, WarpC \cite{warpc}, to estimate dense correspondences across views, and we leveraged our refined maps to generate sparse correspondences. WarpC was trained using an unsupervised method through warp
consistency, and the authors demonstrate state-of-the-art performance and robustness to varying views and domains across image pairs. Our refined maps provide highly accurate relative poses and per-keypoint depth values. Although direct utilization of the matches stored in the maps is possible for correspondence generation, we discovered that the resulting set was too sparse and lacked variation. By reprojecting the keypoints across arbitrary views, we obtained more varied correspondences and better supervision, albeit with more significant occlusion. Lastly, the figure includes iCDC, our method which is detailed in the following subsection.

\subsection{iCDC: Implicit Cross-Domain Correspondences}
\parsection{Generating correspondences with NeRFs.}
A NeRF \cite{mildenhall2021nerf} learns an implicit continuous 3D representation of a scene from a collection of images with known camera viewpoints. In the computer graphics community, NeRFs have been used to synthesize novel photorealistic views of a scene. The computer vision community has adopted NeRFs, exploiting the power of the implicit representations in different domains. \cite{nerfw, hu2023gaia} synthesized data capable of training computer vision models, NeRFs show great promise in the field of scene understanding \cite{wang2022clip}, and they have the potential to revolutionize SLAM algorithms \cite{zhu2022nice, zhu2023nicer}. Beyond synthesizing RGB data, explicit 3D representations of the implicit scenes can be rendered. NeRF-supervision \cite{nerf-supervision} generated correspondences across image pairs, by rendering corresponding depth images and comparing the underlying 3D geometries. The authors generated correspondences across images of an object, using a NeRF trained in a controlled environment with constant lighting conditions, by reprojecting the rendered depth maps from one frame to the other. The authors demonstrated the powerful cross-view information aggregation capabilities of NeRFs, by generating accurate correspondences across reflective objects -- a canonical challenge. 

\begin{samepage}
\parsection{iCDC overview.}
Building on NeRF-supervision \cite{nerf-supervision}, we introduce iCDC, capable of generating accurate correspondences across images captured in different visual domains.
iCDC leverages the fact that the underlying static 3D geometry of a scene, such as buildings or street signs, remains unchanged, regardless of visual domain changes, such as daytime, weather, and seasonal variations.
For each scene, a single NeRF is trained per trajectory, i.e., per visual domain, using the poses from our multi-trajectory refined maps. 
\end{samepage}

Training a single NeRF using data across different trajectories with different visual domains would result in inferior scene representation. NeRF-W \cite{nerfw} addresses this by learning per-image latent appearance codes. We found that this makes NeRFs robust to lighting variations, however, long-term variations are far more challenging to model. Consider the simple example, where a tree that is standing in one trajectory, and has been cut down in another. If a NeRF was trained on both trajectories, it would learn a representation of a scene that can not be correct for both cases and relying on the NeRF's underlying 3D representation of the scene would result in false positive correspondences of the tree. 

Although in iCDC the per-trajectory NeRFs do not share training data, they are expressed in the same coordinate system. Due to the high accuracy of the maps, the implicit 3D representations of the NeRFs share the same space and can be compared to generate real cross-domain correspondences. More specifically, for a pair of images captured along different trajectories, we first render depth maps using the corresponding NeRFs. Subsequently, we reproject the points utilizing the corresponding map poses. 

\parsection{Scaling NeRFs to represent 4Seasons scenes.} 
The 4Seasons dataset is built from long trajectories that span large areas, and representing entire scenes with a single NeRF is challenging. Following recent works \cite{mega-nerf, block-nerf}, we evenly cluster the scenes into blocks based on frame poses and train a small NeRF per block. Since 4Seasons trajectories can revisit the same place multiple times, a NeRF block can contain multiple segments of the same trajectory. Near the edges of the blocks, this setup leads to poor rendering qualities, because there are only a few observations available to train the NeRFs. To overcome this, blocks are expanded by a buffer region. These buffer regions overlap with other blocks, and the frames in these buffer regions are used only for training purposes. Therefore, while the blocks share training data, each frame has a designated NeRF used to render depth maps. For convenience, we refer to the group of NeRFs that represent an individual trajectory as a trajectory-NeRF. At test-time, such a trajectory-NeRF functions exactly like a normal NeRF: frames are rendered at input camera pose locations.

\parsection{Supervising NeRFs with sparse depth.} 
DS-NeRF~\cite{depth-supervision} showed that supervising a NeRF with sparse depth values can significantly improve the rendering performances with fewer training views.
We found that in our scenario, to render accurate depth maps, sparse depth supervision is not only beneficial but necessary. This is because the 4Seasons data is captured using a forward-facing stereo camera, offering minimal view variation. Moreover, the scenes are unbounded, i.e. the scene points lie at large distance variations from the camera. Finally, the environments are uncontrolled, which means that the camera is often exposed to drastic lighting conditions changes along a trajectory. Each of these factors makes interpreting depth from only RGB images a significant challenge. The additional depth supervision provides sufficient information for the NeRFs to effectively learn the underlying 3D representations. 
Following the DS-NeRF approach, we use sparse depth supervision to improve the depth rendering results. We extracted sparse depth values per frame directly from our sparse SfM maps.
 
\parsection{Depth-frame rendering.}
NeRF-Supervision~\cite{nerf-supervision} proposed to treat the rays as depth probability distributions, sampling depth values during runtime. Instead, we follow the Instant-ngp \cite{instant-ngp} implementation. By modeling the transmittance along a ray, we can render depth by estimating the first instance after the light is absorbed by a surface in the scene. Furthermore, we can generate depth data prior to extractor training, instead of having to sample depths during the training.

\parsection{Correspondence generation.}
For a pair of images $\mathcal{I}\in \mathcal{U}$ and $\mathcal{J}\in \mathcal{V}$, belonging to trajectories $\mathcal{U}$ and $\mathcal{V}$ respectively. We render depth maps using trajectory-NeRF-$\mathcal{U}$ and trajectory-NeRF-$\mathcal{V}$. Then, we generate candidate correspondences across the images, by reprojecting the depth map of $\mathcal{I}$ onto $\mathcal{J}$ using the poses stored in the map. In the case where $\mathcal{U}=\mathcal{V}$, the same NeRF will be rendering the depths for both frames.

We use the loop consistency test, similar to WarpC~\cite{warpc}, to filter out invalid correspondences. The candidate pixel of $J$ corresponding to a reference pixel of $\mathcal{I}$, has its own rendered depth value. We repoject this depth value back onto image $I$, arriving at a certain 2D loop consistency distance from the reference pixel. We filter out correspondences with loop consistency distances that are higher than a user-defined loop consistency threshold $\alpha$.

While the loop consistency test removes most of the invalid correspondences, there can be corner cases with false positives. For example, when the relative pose between the reference and query frame is only a transformation along the frame normal (the case between most frame pairs in the 4Seasons dataset), the corresponding center pixels will always pass the loop consistency test. To filter out false positives, we compare the reprojected depth value of a reference pixel, with the depth value of the query pixel, where the absolute difference between the depth values is the depth consistency difference. We filter out correspondences with depth consistency differences that are greater than the depth consistency threshold $\beta$; another user-defined threshold. 

\begin{figure*}[tbh]
    \centering
    \begin{tabular}{@{}c@{}}
    \includegraphics[width=0.9\linewidth]{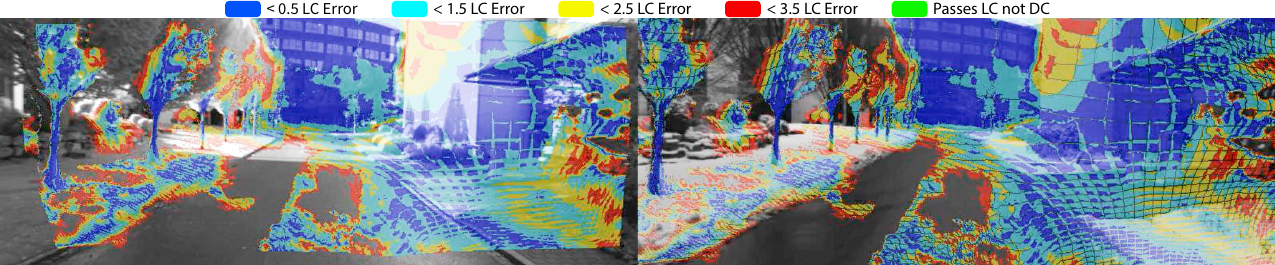} \vspace{-2pt} \\ \vspace{2pt}
    \small (a) Loop Consistency Test\\
    \includegraphics[width=0.9\linewidth]{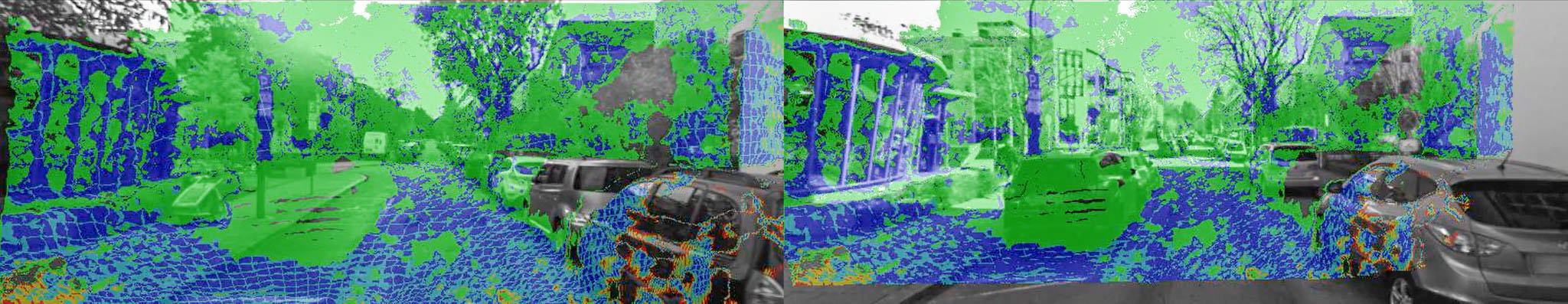} \vspace{-2pt}\\
    \small (b) Loop+Depth Consistency Test \\
  \end{tabular}
  
\caption{\textbf{iCDC Correspondences.} Both figures display image pairs illustrating iCDC Correspondences. Fig. (a) shows pixels 
 of correspondences satisfying loop consistency (LC) tests colored according to LC thresholds. Fig. (b) depicts the same mask, except marking pixels as green that pass the LC test but not the depth consistency (DC) test with a 15cm DC error threshold.}
    \label{fig:correspondence check}
\end{figure*}

We present in Fig. \ref{fig:correspondence check} examples of generated correspondences. In the first example, we filter using only the loop consistency test, while in the second example, we filter using both the loop and the depth consistency tests.

\section{Experimental Setup Details}
\label{sec:exp}

\subsection{Train-test-split}
We train on two scenes of the 4Seasons dataset: Office Loop and Neighborhood and test on the third scene Business Campus. 
Both in training and testing splits, we employ spatiotemporal keyframing to minimize image pairs with minor variations. In the training split, along a single loop of a trajectory, consecutive reference frames are keyframed approximately 6 meters apart. For each reference frame, we sample up to two spaced-out query frames per trajectory loop, with maximum distances of 6 meters from the reference. If the trajectory loops around, the number of reference keyframes and the number of query frames per reference are both multiplied by the number of loops. This methodology yields 16266 intra-domain and 77428 cross-domain training pairs. We explain the keyframing of the testing data for each experiment.

\subsection{Constructed Maps}
We outlined our approach in Sec. \ref{sec:dataset} for constructing joint-trajectory maps, to generate accurate relative poses across domains. While we have constructed global maps for scenes, to achieve finer refinements, we broke the problem down into multiple subproblems. As we only ever compare image pairs, we only need relative poses across two trajectories at a time. Therefore, for a given scene, we constructed maps for all trajectory pairs. Furthermore, for each trajectory pair, we construct two types of maps. For benchmarking local features, we construct maps consisting of a subset of the frames available in each trajectory (the frames with registered geolocations). On the other hand, for NeRF training, the more frames the better. Therefore, we generate a second set of maps using all available frames, as described in Sec. \ref{sec:dataset}.

\subsection{iCDC Setup}
\parsection{NeRF training.}
While our multi-trajectory maps were built using only images captured by a single camera, the 4Seasons dataset provides stereo image pairs. To train the NeRFs, we use images captured from both cameras.
For efficient training of the NeRFs, we use Instant-ngp, as it is designed for fast processing, ensuring quick training and rendering crucial for our block-wise NeRF approach. The scene is divided such that the farthest apart frames in each block are approximately 64 meters apart. For more details about the engineering solutions necessary for representing unbounded outdoor scenes using Instant-ngp, check out our supplementary material.

\parsection{Generating correspondences.}
We generate correspondences across an image pair using the corresponding rendered depth maps, then filter out invalid correspondences through loop consistency and depth consistency tests, as detailed in section \ref{sec:supervision}. For the loop consistency test, we set a loop consistency threshold of $\alpha=2$ (in pixel units). For the depth consistency test, anchoring poses using geolocations allowed us to use metric units to set our depth consistency threshold. We observed a meaningful improvement in performance with the implementation of the depth consistency test. We selected a threshold of $\beta=15\text{ cm}$.

\subsection{Extractor Training Setup}
While our cross-domain correspondences can supervise any feature extractor, R2D2 \cite{revaud2019r2d2} is the primary network of our investigation. Our choice of R2D2 was influenced by its unique training loss characteristics. Besides optimizing for keypoint repeatability, which is further divided into ``peakiness'' and ``cosimilarity'' losses, it also optimizes for descriptor reliability. Each of these losses can be explored, providing an avenue for thorough investigation. In fact, we found that the repeatability loss needed to be adapted for cross-domain correspondence supervision. The loss function was designed for homography-based correspondences, where there are no significant geometrical variations across the matched keypoints. For cross-domain correspondences, we found that the loss needed to be adapted. 
To train R2D2, for all cross-domain correspondence methods, we employed the cosimilarity loss supervision exclusively for intra-domain training pairs. For more details and insights about the losses, please see our supplementary material. 

To demonstrate the importance of real cross-domain correspondences, we trained an additional extractor to demonstrate that the improvements were not network-specific. We train SiLK \cite{gleize2023silk} -- we selected it due to the simplicity of its training pipeline. 
We trained both networks using their default pipelines, making only the requisite adjustments to accommodate our correspondences. In the case of R2D2, we updated the losses, and in the case of SiLK, the pipeline needed to be updated to support non-homography-based correspondences.

In both the R2D2 and SiLK training pipelines, images are augmented using noise, as well as random cropping and scaling. Additionally, when supervising with real correspondences, we apply homographic transformations as a form of data augmentation. Unlike the SiLK pipeline which only supports homography-based supervision, R2D2 has a function to map ground truth correspondences across an image pair post-application of a perspective homographic transformation to the query image. We adopt this strategy from R2D2 to adapt the SiLK pipeline, enabling it to utilize real correspondences.

To speed up training, we initialize both R2D2 and SiLK with their provided default weights. However, we use a large enough learning rate so that the previous patterns are mostly overwritten. Initializing using the default weights only marginally improves over initializing with random weights.

\section{Results}
\subsection{Overview of Experiments}
We train R2D2 using multiple correspondence methods on the 4Seasons dataset and observe a significant reduction in the performance gap when utilizing real cross-domain correspondences. To ensure that the performance improvements are not solely attributed to the training data, we establish two baselines for our networks. Firstly, we use the default weights as a baseline. Secondly, we create another baseline by training R2D2 using its default pipeline (homographies) on the 4Seasons dataset. In our first experiment, we train using two straightforward real cross-domain correspondence generation methods. We introduced iCDC to enhance the accuracy of the cross-domain correspondences. In a second experiment, we benchmark the iCDC-supervised extractor.
For that, we generate correspondences using different methods and use them to train our feature extraction and description backbone. 

We summarize the correspondence generation methods:
\begin{itemize}
    \item \textbf{Homography:}  \ The default R2D2 training scheme is utilized. Individual images are loaded, a perspective homography is applied, and the resulting image pairs are then individually augmented.
    \item \textbf{Dense Correspondence Network:} \ We use WarpC \cite{warpc} to estimate dense correspondences between an image pair. Image pairs are loaded, and WarpC estimates are used as ground truth. To filter out inaccuracies, a loop consistency test is used. Both images are individually augmented, one of which is also augmented using homographic transformation. This same augmentation strategy was also implemented for the following two methods. 
    \item \textbf{Sparse SfM structure:} \ We extract sparse depth values from our maps, and reproject them across image pairs to generate correspondences. 
    \item \textbf{iCDC:} \ Per training image pair, depth maps, rendered by corresponding trajectory-NeRFs, are used to generate correspondences. Candidate correspondences are filtered using loop and depth consistency tests. 
\end{itemize}

\subsection{Qualitative Correspondence Analysis}
\label{subsec:exp qualitative}
\begin{figure*}[t!]
    \hspace{-7pt}
    \centering
    \setlength\tabcolsep{1pt}
    \begin{tabular}{@{}r@{}}
   \includegraphics[width=0.84\linewidth]{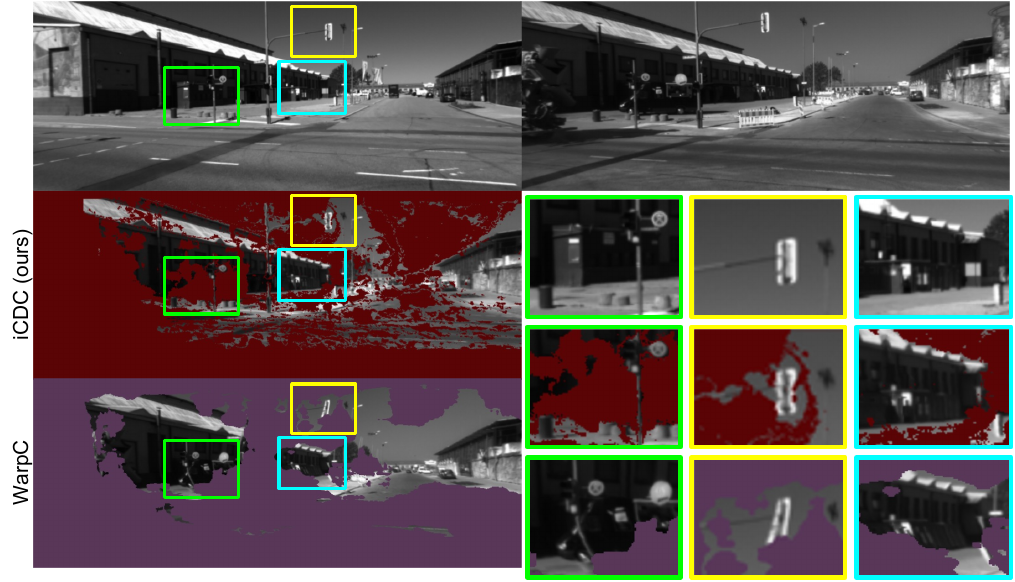} \\
  \end{tabular}
    \caption{\textbf{Comparison Between Our NeRF Correspondences and WarpC~\cite{warpc} Correspondences.} Correspondences are generated for an image pair (top row). The left column's second and third rows show WarpC and iCDC correspondences respectively. More specifically, the right-hand side image is warped onto the left-hand side image following the correspondences. We zoom in on windows for detailed examination. In the case of accurate correspondences, the warped frames align with the left-hand side frame.}
    \label{fig: warpc vs nerf corresp}
\end{figure*}

In Fig. \ref{fig: warpc vs nerf corresp}, we compare the quality of correspondences between the dense correspondences estimator WarpC and iCDC. For a fair comparison, we filter out candidate correspondences (red and purple regions) using only the loop consistency test for both methods. We adjusted the thresholds for both methods to generate a similar number of correspondences per image pair. For an image pair (top row), we generate correspondences using both methods. Using the correspondences, we warp the right-hand side image onto the left-hand side image. Effectively, we use the correspondences to align the two images. The warped images are in the second (iCDC) and third (WarpC) rows, with windowed regions zoomed in. 

For both methods, correspondence errors are small-scale and are often visually hard to identify, however, WarpC had significantly more frequent failure cases. In the zoomed-in regions, the warped dense correspondence estimator frames are significantly misaligned. 

\subsection{Relative Pose Estimation on Our Benchmark}
\begin{figure*}[t!]
    \centering
    \begin{tabular}{@{}cc@{}}
    \begin{minipage}{0.5\textwidth}
            \centering
            \hspace*{30pt}\includegraphics[height=7.5cm]{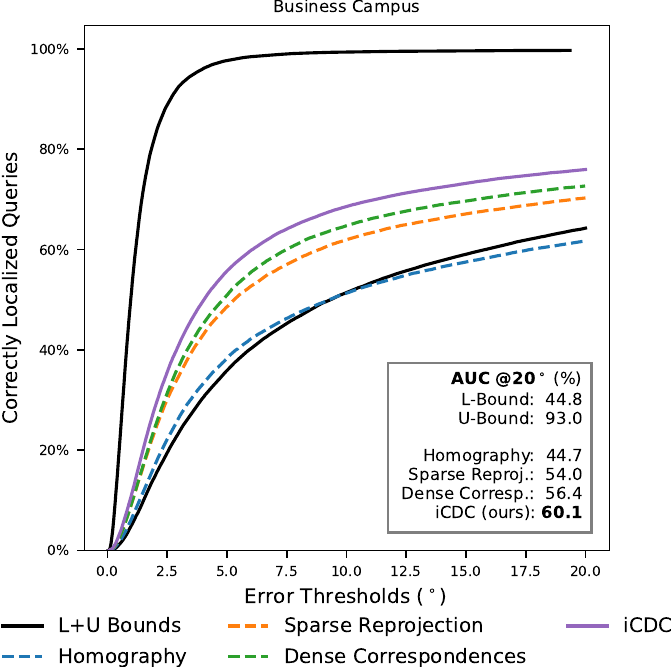} \\
            \small (a) Supervising R2D2
        \end{minipage} &
        \begin{minipage}{0.5\textwidth}
            \centering
            \hspace*{-40pt}\includegraphics[height=7.5cm]{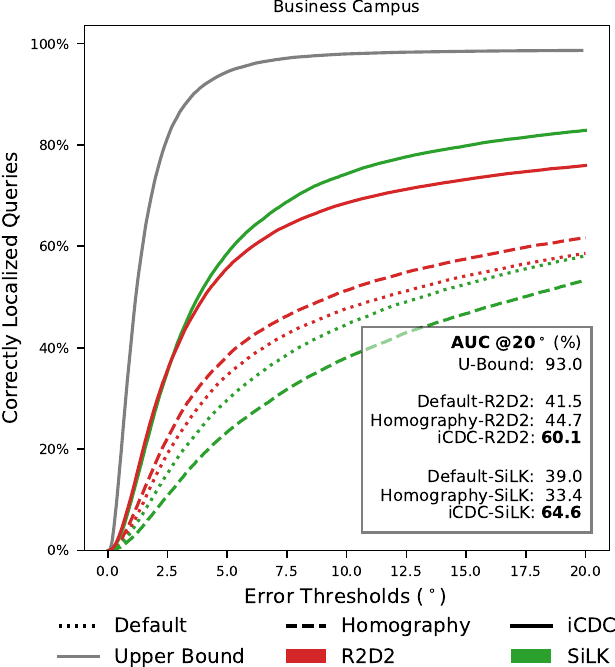} \\
            \centering
            \small (b) Extending supervision to SiLK
        \end{minipage}
  \end{tabular}
   \vspace{0pt}
    \caption{\textbf{Cumulative Distribution of Cross-domain Pose Errors on the 4Seasons Dataset \cite{4seasons}.} Our trained R2D2 extractors are benchmarked using our ground truth poses (a). Lower and upper bounds (L+U Bounds) are the error distributions of the best cross-domain localization method (SuperPoint~\cite{superpoint}) and R2D2~\cite{revaud2019r2d2} evaluated on intra-domain respectively. Dashed lines are our trained extractors used to benchmark iCDC. Furthermore, we compare our trained SiLK performances against R2D2 performances (b).}
    \label{fig: business campus relative AUC}
\end{figure*}

For relative pose estimation, we apply our benchmark setup described in section \ref{sec:benchmarking}, evaluating only on Business Campus trajectories. We choose reference frames at 10m increments along the trajectories. Query frames are selected at two-meter increments, a maximum of eight meters from the reference frame. Using extractors, we estimate relative poses between image pairs, evaluating performances using the ground truth poses in our maps. For a comprehensive understanding of error distributions in cross-domain localization, we plot the cumulative error curves in Fig. \ref{fig: business campus relative AUC}.

\parsection{Discussing primary extractor results (R2D2).}
In Fig. \ref{fig: business campus relative AUC} (a), we set SuperPoint with default weights as the lower bound -- the best-performing pre-trained extractor on cross-domain localization. As an upper bound, we set the intra-domain localization performance of R2D2 with default weights. 

Training R2D2 using their default pipeline (homography) does improve the cross-domain localization performance on our benchmark, however, the improvement is only marginal. It is still outperformed by SuperPoint with default weights, indicating that the meaningful improvements are not attributed to the 4Seasons training images. Employing supervision using cross-domain correspondences uniformly achieves at least approximately a 50\% reduction of the median error on cross-domain localization. A significant performance gap is observed between the homography-based method and our worst-performing extractor supervised by cross-domain correspondences. This notable performance increase highlights the need for data-centric approaches. Even naive methods, like supervising using sparse map-based correspondences or employing estimates from dense correspondence networks, can already meaningfully enhance robustness against long-term variations.

Despite the high accuracy of the correspondences extracted from the sparse maps, using the estimates from the dense correspondence network yields better results. Although the accuracy of correspondences is crucial, the high variation of positive anchors during training, provided by the dense correspondences, was also an important factor. Through our qualitative analysis, we identified that generating correspondences using our iCDC network results in higher accuracy dense correspondences compared to the dense correspondence estimator WarpC. We achieved the best results when supervising R2D2 using iCDC correspondences, due to the increased density and improved accuracy of the correspondences. Referring to the AUC of the pose errors at a threshold of 20°, we improved the cross-domain localization performance of R2D2 by 44.8\% and reduced the performance gap between the default R2D2 and R2D2 trained using our iCDC method by 36.1\%. 

\parsection{Extending discussion to SiLK.}
In order to validate that the findings of this work are not confined to R2D2, we extended our training to SiLK using a smaller subset of the correspondence methods. For clarity, we have separated the results of SiLK supervision in a separate figure (Fig. (b)). SiLK exhibits some distinct characteristics. As noted in the benchmarking section (Sec. \ref{sec:benchmarking}), SiLK achieves state-of-the-art performances within the same domain localization, however, it achieves the worst performance in cross-domain localization. Our experiments indicate that SiLK fits training data more aggressively than R2D2, and if the training data is not varied enough, or is not visually similar enough to the test domain, then SiLK does not generalize well.

We trained on the 4Seasons data utilizing the default SiLK training pipeline (homography). In contrast to the observation with R2D2, the resulting network underperforms compared to SiLK with default weights. Overfitting to the training scenes, with the 4Seasons dataset being less varied than SiLK's default training data, yields the worst cross-domain performance in this work. Interestingly, however, supervising SiLK with iCDC correspondences leads to the best performance across all our networks. The aggressive fitting of cross-domain correspondences proves highly effective on our test scene, despite the limited training image variation.

\subsection{Absolute Pose Estimation on Our Benchmark}
We localize a query image using feature extractors with respect to a reference 3D map. Our approach follows the benchmark setup for absolute pose estimation as previously detailed in sec. \ref{sec:benchmarking}. For each trajectory in the testing scene, we triangulate an SfM map using an extractor and ground truth poses. For image retrieval, we utilize NetVLAD~\cite{arandjelovic2016netvlad}. The localization results under thresholds (0.25m, 2$^\circ$), (0.5m, 5$^\circ$), and (5m, 10$^\circ$) are reported in Tab. \ref{tab: business campus absolute pose AUC}. Metrics are averaged per reference trajectory and are also globally averaged. 

\begin{table}[ht]
\setlength\tabcolsep{4.4pt}
  \center
    \caption{\textbf{Cross-Domain Absolute Pose Accuracies on the 4Seasons Dataset \cite{4seasons}.} For our test scene, the percentages of accurately localized query images under thresholds (0.25m, 2$^\circ$), (0.5m, 5$^\circ$), and (5m, 10$^\circ$) are reported, evaluated against our ground truth poses. Results using default weights, our trained R2D2 networks, and our trained SiLK networks are grouped separately.}

      \scriptsize
     \begin{tabular}{@{}rccca@{}}\toprule
 
\textbf{Traj.}&BC1 & BC2 & BC3 & ALL\\ \midrule
\multicolumn{5}{l}{\textbf{Pretrained Benchmarks}}\\
SuperPoint~\cite{superpoint}  & 61.4/78.1/96.3 & 67.1/79.7/98.4 & 65.1/81.9/97.6 & 64.5/79.9/97.4 \\
R2D2~\cite{revaud2019r2d2}          &   55.0/75.2/97.6 & 63.6/79.3/99.1 & 61.9/81.2/98.0 & 60.2/78.5/98.2\\
D2-Net~\cite{dusmanu2019d2}        &  57.4/79.6/98.0 & 60.7/80.5/99.2 & 60.9/82.4/98.6 & 59.7/80.9/98.6\\
SiLK~\cite{gleize2023silk}      & 34.0/46.9/70.1 & 50.2/64.9/88.3 & 42.5/60.5/82.2 & 42.2/57.4/80.2 \\
\midrule
\multicolumn{5}{l}{\textbf{Supervising R2D2 \cite{revaud2019r2d2}}}\\
Homography          &   61.6/77.5/97.9 & 67.3/80.1/99.3 & 64.4/82.6/98.2 & 64.4/80.1/98.5\\
 Sparse Map  & 64.2/79.2/\textbf{98.4} & \textbf{70.7}/81.1/99.5 & 67.5/84.0/98.5 & 67.5/81.4/98.8 \\
Dense Corresp.        &   64.4/\textbf{80.8}/98.3 & 67.4/\textbf{82.0}/99.5 & 66.8/\textbf{85.4}/98.7 & 66.2/82.7/98.8 \\
iCDC        &  \textbf{66.3}/80.6/\textbf{98.4} & 69.1/81.8/\textbf{99.6} & \textbf{69.0}/85.3/\textbf{98.8} & \textbf{68.2}/\textbf{82.6}/\textbf{98.9}\\
\midrule
\midrule
\multicolumn{5}{l}{\textbf{Supervising SiLK \cite{gleize2023silk}}}\\
Homography           &   25.7/38.9/64.1 & 42.3/58.4/83.1 & 41.7/59.0/81.1 & 36.6/52.1/76.1\\
iCDC        &  63.0/77.2/97.1 & 69.2/80.1/98.5 & 67.4/83.7/98.0 & 66.5/80.3/97.8\\
\bottomrule
\end{tabular}

\label{tab: business campus absolute pose AUC}
\end{table}

We observe similar patterns as in the relative pose estimation case, both for R2D2 and SiLK. Supervising R2D2 using the default pipeline increases the cross-domain localization performance, however, does not outperform SuperPoint with default weights. Supervising R2D2 using cross-domain correspondences uniformly outperforms the default extractors, and moreover, the accuracy of our iCDC correspondences yields the best results. However, the reference maps contain a very dense set of frames. Consecutive frames are no more than 30 cm apart, and the Business Campus trajectories loop around three times. Therefore, the task of NetVLAD is made easy, and retrieved reference and query frames have very small relative poses, simplifying the localization task. We find that there is less room for improvement than in the relative pose estimation case, with the performance increases being less dramatic. In relative pose estimation, on the other hand, we engineer a broader range of relative poses between test image pairs. 

\subsection{Evaluation of Our Correspondences on the Visual Localization Benchmark}
\label{subsec:exp longloc benchmark}

To investigate the generalizability of our feature extractors supervised by iCDC, we evaluate their localization performances in the long-term localization benchmark \cite{sattler2018benchmarking}. The benchmark datasets are captured using different camera setups, with each dataset presenting its own set of challenges. 

\parsection{Overview of the benchmark datasets.}
CMU-Seasons \cite{cmu} is the setup most similar to our own: a camera mounted on a car with scenes ranging across rural and urban and car trajectories ranging across visual domains. However, while the 4Seasons cameras are forward-facing, the CMU cameras are rotated 45$^\circ$. Furthermore, a significantly larger portion of the CMU dataset is looking at vegetation than in the 4Seasons dataset. 

While the RobotCar-Seasons dataset \cite{robotcar} has images captured by sidewards facing cameras, it also has rear-view images which are more similar to the 4Seasons scenario. In fact, the query frames in the benchmark are all rear-view images. Furthermore, RobotCar is captured in an urban scene, with more recognizable geometries than CMU. However, in addition to visual domain variations similar to those of 4Seasons, one of the key benchmarks of RobotCar is night-time image localization against a day-time reference map. We do not have night-time images in our training set. Moreover, the image qualities are significantly lower than those of 4Seasons, in particular, the night-time images with strong motion blur. 

Aachen Day-Night \cite{aachendn}  also benchmarks night-time image localization and the quality of those images is much higher. However, Aachen Day-Night presents the greatest challenge for our extractors. It is a hand-held camera dataset that presents drastically different relative views between images, and while the 4Seasons images mostly look along roads with buildings along the sides, the Aachen Day-Night images mostly look directly at buildings. Localization results, using NetVLAD for image retrieval, on the long-term visual localization benchmark are listed in Tab. \ref{tab:benchmarking long term loc}.

\begin{table*}[ht]
\setlength\tabcolsep{6.2pt}
  \center
        \caption{\textbf{Benchmarking Long-term Visual Localization.}We evaluate the performance of our extractors on the long-term localization benchmark \cite{sattler2018benchmarking}, and compare them to the performances of SuperPoint, R2D2, and SiLK with default weights. CMU Seasons and RobotCar-Seasons are autonomous driving datasets, similar to our training data. Aachen Day-Night presents the greatest challenge due to the large domain variation between hand-held and car-mounted camera setups.}

      \scriptsize
 \begin{tabular}{@{}rcccccccccc@{}}\toprule
\textbf{Scenes}& \multicolumn{3}{c}{CMU Seasons Extended~\cite{cmu}} & \phantom{} & \multicolumn{2}{c}{RobotCar-Seasons v2 \cite{robotcar}} & \phantom{} & \multicolumn{2}{c}{Aachen Day-Night v1.1~\cite{aachendn}} \\
\cmidrule{2-4} \cmidrule{6-7} \cmidrule{9-10}
\textbf{Traj.}& Urban & Suburban & Park && Day All & Night All && Day & Night \\ \midrule
SuperPoint~\cite{superpoint} & 93.5 / 96.1 / 98.2  & 84.2 / 86.9 / 92.9  & 78.2 / 81.9 / 85.8  && 64.2 / 93.7 / 99.1 & 14.0 / 27.7 / 33.8 && 87.1 / 93.3 / 96.7 & \textbf{68.1} / 80.6 / 90.1 \\
R2D2~\cite{revaud2019r2d2} & 94.4 / 97.3 / 99.0 & 85.3 / 88.9 / 93.5 & 74.2 / 78.8 / 83.6  && 64.5 / 94.4 / 99.2 & 16.8 / 32.6 / 39.4 && \textbf{88.0} / \textbf{95.3} / \textbf{97.9} & 67.0 / \textbf{86.4} / \textbf{96.3} \\
iCDC-R2D2 & \textbf{96.5} / \textbf{98.8} / \textbf{99.4} & \textbf{93.4} / \textbf{94.9} / \textbf{97.1} & \textbf{87.5} / \textbf{90.3} / \textbf{92.9} && \textbf{64.9} / \textbf{94.6} / \textbf{99.4} & \textbf{24.5} / \textbf{45.5} / \textbf{52.2} && 87.7 / 93.7 / 96.6 & 67.5 / 84.3 / 93.7 \\
\midrule
\midrule
SiLK \cite{gleize2023silk} & 57.6 / 62.0 / 68.0 & 42.5 / 47.9 / 57.6 & 17.0 / 19.6 / 26.6 && 62.2 / 90.8 / 96.1 & 12.6 / 18.4 / 21.7 && 78.8 / 84.3 / 89.8 & 38.2 / 48.2 / 55.5 \\
iCDC-SiLK & 92.2 / 94.9 / 96.9 & 81.8 / 84.9 / 89.2 & 63.2 / 67.2 / 72.4 && 64.5 / 94.0 / 99.2 & 15.4 / 26.8 / 34.3 && 83.9 / 90.0 / 94.1 & 60.7 / 76.4 / 85.3 \\
\bottomrule
\end{tabular}
\label{tab:benchmarking long term loc}
\end{table*}

\parsection{Discussing primary extractor results (R2D2).}
On the driving datasets, iCDC-R2D2 correspondences outperformed default SuperPoint and R2D2 in all metrics. In CMU Seasons, there are major performance increases in both the Suburban and Park scenes. In the Urban scenes, the performance increase is less drastic, however, given the already very high performances, the increase is meaningful. In RobotCar-Seasons day-time localization, our extractor only marginally outperforms the baselines. This might be a NetVLAD bottleneck, or due to the lower image qualities. In night-time localization, however, our extractor significantly outperforms the other baselines, suggesting that supervising domain variation improves performances in unseen domains. These results underscore the importance of improved data-centric approaches. Despite the lack of view variation in the 4Seasons dataset, we see significant performance increases in long-term localization scenarios. The real-world variations across correspondences even improve localization performances in unseen visual domains, such as in night-time localization.  

Despite the large view-point domain shift, our extractor achieves competitive results on the Aachen Day-Night benchmark, outperforming SuperPoint in almost all metrics. R2D2 with default weights achieved the best results in most metrics, however, it was trained on Aachen Day-Night day-time data, as well as on day-to-night style-transferred versions of the same images. 
For detailed CMU and RobotCar results, please see the supplementary material. 

\begin{samepage}
\parsection{Extending the discussion to SiLK.}
Our benchmark revealed that SiLK aggressively fits the training data, struggles with generalizing, and with default weights achieves low cross-domain localization performances. Furthermore, training on 4Seasons with their default pipeline reduces the performance by overfitting. On the other hand, we found that supervising SiLK with iCDC correspondences yielded the best results on our benchmark. That being said, generalizing to the long-term localization benchmarks is a separate challenge. SiLK with default weights performs poorly on CMU and Aachen Day-Night, although it has comparable results to the other baselines on RobotCar. Similarly to the case of our benchmark, iCDC supervision significantly improves the performance of SiLK on the long-term localization benchmarks by aggressively fitting the domain variations. This is only enough to make certain metrics comparable in CMU and Aachen. As RobotCar-Seasons' rear-view images present the scenario most similar to that of 4Seasons, iCDC-SiLK achieves competitive results.   
\end{samepage}
\section{Conclusion}
\label{sec:conclusion}
Modern feature extraction methods advance the state of the art by developing more powerful network architectures and novel optimization schemes. They supervise their networks by applying perspective homographies and augmenting data to generate training tuples, enabling the training schemes to run flexibly in a self-supervised manner. However, in long-term localization scenarios, visual and perspective variations arise that are beyond the simulation capabilities of the commonly used data augmentation practices. This creates a bottleneck that model-centric approaches cannot overcome. To achieve successful life-long deployment, a shift towards improved data-centric approaches is imperative.

We established a novel benchmark for the long-term visual localization scenario, enabling flexible evaluation of visual localization performances across a broad spectrum of visual domains. The 4Seasons dataset offers trajectories of images captured in several kinds of scenes under varied weather conditions, seasons, and daytime lighting conditions. Through joint trajectory SfM map refinement, we generated highly accurate maps, providing ground truth relative poses between images captured under different visual domains. We thoroughly evaluated the performances of state-of-the-art extractors on our benchmark, revealing a large performance gap between intra- and cross-domain localization.

To achieve accurate cross-domain localization, real cross-domain correspondences are needed to supervise networks. To accentuate this point, we implement readily available methods to generate correspondences across domains. We train two feature extractors, R2D2 and SiLK, using these correspondences, and significantly reduce the performance gap. We propose a novel data-centric approach to further improve cross-domain localization performances. Implicit Cross-Domain Correspondences (iCDC) is a method to generate accurate correspondences across visual domains by comparing the underlying implicit 3D geometries of domain-specific NeRFs. By supervising networks with real cross-domain correspondences, we reduced the performance gap by 36.1\%, as evidenced in our relative pose estimation benchmark.

This work represents a significant step forward in enhancing the robustness and reliability of visual localization pipelines for life-long deployment. It highlights the need for data-centric methods in improving cross-domain localization performance. Future works could explore other data-centric approaches and could refine the efficiency and scalability of the proposed method. The creation of easily deployable cross-domain correspondence generation pipelines, robust against dynamic objects within scenes, would enable scaling to larger amounts and more varied datasets. This advancement, in turn, holds potential for training networks capable of achieving intra-domain localization performances in long-term deployments. The insights gained from this research provide a solid foundation for subsequent studies focused on optimizing the performance of feature extraction networks.
\clearpage
\appendix
\setcounter{figure}{0}  
\setcounter{table}{0}  
\renewcommand{\thetable}{A.\arabic{table}}
\renewcommand{\thefigure}{A.\arabic{figure}}
\subsection{Training Learned Matchers Using iCDC}
We additionally explored two related fields that are relevant to this work. (i) NeRFs can be used to generate data. Instead of storing a dataset and sampling from a discrete set of frames, NeRFs could be used to flexibly sample images along a continuous range of camera poses. Furthermore, views can be extrapolated from the original camera trajectory, introducing novel views that have the potential to improve the generalization of networks. Moreover, (ii) learned matchers \cite{superglue, lindenberger2023lightglue} rely on local features and learn matching patterns across image pairs. They have significantly increased localization performances in complex environments, such as day-night localization in the long-term localization benchmark \cite{sattler2018benchmarking}. We study these two extensions of iCDC.

\parsection{Training with novel views.}
We explore the impact of using rendered images in training. In particular, we were interested in two factors: (1) can we maintain performances when training on rendered data, and (2) can we improve generalization to other datasets by using extrapolated views? To this end, we generate a secondary dataset rendered by our NeRFs. The images are rendered at the same XYZ coordinates as the training data, except we incrementally rotated the views around the gravity axis left and right. We found, however, that because of the forward-facing setup, extrapolation was very difficult. We limited the rotation to $\pm \frac{\pi}{8}$, and even then, there was a noticeable decrease in rendering quality. For training, we used iCDC correspondences.

After training, we evaluated on our benchmark. It is important to note that the extrapolated views will not improve performance on our benchmark, because such variations are not present. We measured the cross-domain performance; pose AUC in \% under thresholds $5^\circ / 10^\circ / 20^\circ$ is: 28.1/43.1/55.7. The performance dips in comparison to our best network iCDC-R2D2 (31.5/47.4/60.1), but it did outperform all other feature extractors trained using homography (the best being 20.2/32.7/44.7), and, while being outperformed, achieved comparable results to the best performing naive cross-domain correspondence generation approach, the dense correspondence network supervision: 28.1/43.6/56.4. We present long-term localization benchmark results for many more extractors in Tab. \ref{tab:appendix long-term loc benchmark}. We get more balanced results for the training with synthesized views (Nov.-iCDC-R2D2+NN). Thanks to the improved generalizability, it performs approximately as well as iCDC-R2D2, and outperforms it in RobotCar and Aachen Day-Night in many metrics, despite the suboptimal quality of the training data.

\parsection{Learned matchers.}
LightGlue \cite{lindenberger2023lightglue} was recently introduced, improving the trainability, efficiency, and performance of the deep keypoint matcher SuperGlue \cite{superglue}. We train a vast array of networks using the default setup, with two key adaptations. (a) add new types of augmentation to data in the training pipeline, and (b) train on the 4Seasons dataset using iCDC. (a) is important because we found that, while feature extractors overfit to texture patterns in images, learned matchers overfit to repetitive matching patterns across image pairs. The default LightGlue training pipeline did not apply homographic transformations or cropping to the training pairs when training on Megadepth \cite{li2018megadepth}, as the types of images are much more varied than in our case. In the 4Seasons dataset, it is very easy to overfit to matching patterns. For example, keypoints on the left will stay on the left in almost all cases, because the car moves forward in a straight line. While overfitting to matching patterns on the 4Seasons dataset leads to the best performances on our benchmark, the resulting matchers can not generalize at all to other datasets. We introduced a small amount of random cropping to explore the effect of overfitting to matching patterns. We only present these results. Furthermore, in a different approach, we also use our rendered extrapolated views for training in an attempt to generalize to other datasets. We only present results for training LightGlue with novel views on the long-term localization benchmark.

\begin{figure}[t!]

 \centering
  \includegraphics[width=0.85\linewidth]{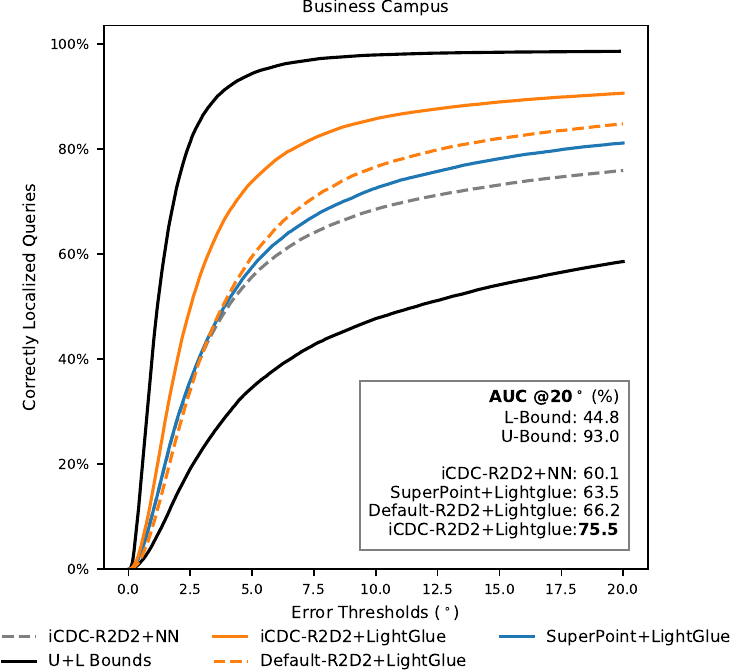}
\caption{\textbf{Cumulative Distribution of Cross-domain Pose Errors on Our Benchmark Dataset.} Keypoints are matched using either Euclidean distance-based Nearest Neighbors (NN), or using LightGlue \cite{lindenberger2023lightglue}. The lower bound is the SuperPoint \cite{superpoint} with default weights. The upper bound is the intra-domain localization performance of \cite{revaud2019r2d2} with default weights.}
\label{fig: appendix lightglue AUC}
\end{figure}

In Fig. \ref{fig: appendix lightglue AUC}, we present a small subset of the networks we trained, benchmarking against our extractors trained with iCDC using Euclidean distance-based nearest neighbors (NN) matching, and against pretrained SuperPoint+LightGlue. As expected, matching with LightGlue led to a huge increase in performance over SuperPoint+NN, the lower bound in the plot. 
While SuperPoint+LightGlue outperforms our network with NN matching, iCDC-R2D2+NN, the results are comparable, highlighting the importance of cross-domain correspondence supervision.

To investigate the importance of descriptor quality when matching with learned matchers, we first trained LightGlue with default-R2D2 features. It marginally outperforms SuperPoint+LightGlue on our benchmark, however, to draw any conclusions about the networks or the training strategy, a further ablation study would be needed. Subsequently, we trained LightGlue using iCDC-R2D2. The resulting network and matcher achieved the greatest reduction of the performance gap with respect to default-R2D2, 66.0\% when considering AUC @20$^\circ$), and outperformed SuperPoint+LightGlue by 18.9\%. On top of highlighting the power of learned matchers, the large gap between default-R2D2 and iCDC-R2D2 using LightGlue underscores the importance of data-centric research for training local features. 

\begin{table*}[ht]
\setlength\tabcolsep{3.2pt}
  \center
        \caption{\textbf{Benchmarking Long-term Visual Localization.} Performances of network matcher pairs are compared. NN means euclidean distance-based nearest neighbor matching. LGlue refers to LightGlue \cite{lindenberger2023lightglue}. Nov. means the network was trained using rendered novel views of our NeRFs.}
      \scriptsize
 \begin{tabular}{@{}rcccccccccc@{}}\toprule
\textbf{Scenes}& \multicolumn{3}{c}{CMU Seasons Extended~\cite{cmu}} & \phantom{} & \multicolumn{2}{c}{RobotCar-Seasons v2 \cite{robotcar}} & \phantom{} & \multicolumn{2}{c}{Aachen Day-Night v1.1~\cite{aachendn}} \\
\cmidrule{2-4} \cmidrule{6-7} \cmidrule{9-10}
\textbf{Traj.}& Urban & Suburban & Park && Day All & Night All && Day & Night \\ \midrule
SuperPoint+LGlue & 	96.1 / 98.8 / 99.6 & 93.3 / 96.0 / 97.9 & 89.1 / 92.7 / 95.0 && 63.9 / 93.8 / 99.3 & 24.5 / 45.9 / 56.6 && 	90.0 / 96.0 / 99.3 & 75.9 / 90.6 / 98.4 \\
\midrule
\midrule
R2D2+NN & 94.4 / 97.3 / 99.0 & 85.3 / 88.9 / 93.5 & 74.2 / 78.8 / 83.6  && 64.5 / 94.4 / 99.2 & 16.8 / 32.6 / 39.4 && \textbf{88.0} / \textbf{95.3} / \textbf{97.9} & 67.0 / \textbf{86.4} / \textbf{96.3} \\
iCDC-R2D2+NN & \textbf{96.5} / \textbf{98.8} / \textbf{99.4} & \textbf{93.4} / \textbf{94.9} / \textbf{97.1} & \textbf{87.5} / \textbf{90.3} / \textbf{92.9} && 64.9 / 94.6 / \textbf{99.4} & \textbf{24.5} / 45.5 / 52.2 && 87.7 / 93.7 / 96.6 & 67.5 / 84.3 / 93.7 \\
Nov.-iCDC-R2D2+NN & 	96.5 / 98.7 / 99.4 & 92.8 / 94.3 / 96.7 & 85.9 / 89.0 / 91.7 && 	\textbf{65.1} / \textbf{94.7} / \textbf{99.4} & 23.5 / \textbf{47.1} / \textbf{52.7} && 86.9 / 93.7 / 96.7 & \textbf{68.1} / 84.3 / 93.7 \\
Default-R2D2+(iCDC-LGlue) & 	86.4 / 90.5 / 94.9 & 77.7 / 83.5 / 92.3 & 53.2 / 60.5 / 74.5 && 	63.9 / 93.8 / 99.0 & 5.6 / 12.1 / 29.1 && 70.3 / 81.1 / 89.3 & 28.3 / 39.3 / 59.7 \\
iCDC-R2D2+(iCDC-LGlue) & 	90.9 / 94.1 / 97.3 & 87.9 / 91.8 / 96.5 & 74.0 / 80.0 / 88.6 && 64.3 / 94.2 / 99.4 & 13.5 / 23.5 / 39.4 && 	67.1 / 76.6 / 86.4 &  27.7 / 39.8 / 62.3 \\
Nov.-iCDC-R2D2+(iCDC-LGlue) & 	89.8 / 93.5 / 97.3 & 86.0 / 90.3 / 95.8 & 71.8 / 78.6 / 88.3 && 	64.4 / 94.6 / 99.6 & 	16.8 / 31.9 / 49.2 && 69.8 / 80.6 / 88.7 & 	37.2 / 50.8 / 71.7 \\
\bottomrule
\end{tabular}
\label{tab:appendix long-term loc benchmark}
\end{table*}

Using the same setup, i.e. same camera mounted on a car facing forward, these networks and matchers are close to closing the gap even when generalizing to new scenes. This is achieved, on one hand thanks to the features, but on the other hand by overfitting to the matching patterns, and therefore, can not generalize well to new setups. We highlight this by providing an ablation study on the long-term localization benchmark in Tab. \ref{tab:appendix long-term loc benchmark}. SupPoint+LightGlue outperforms all other network-matcher pairs, with only iCDC-R2D2+NN being competitive in all datasets except Aachen Day-Night. Without random cropping in the LightGlue training, the performances were significantly worse, and introducing cropping only makes the results comparable but not competitive, except on RobotCar day-time data, because RobotCars rear-view camera has similar characteristics to the 4Seasons setup. Similarly to our benchmark, training LightGlue using iCDC-R2D2 instead of default-R2D2 yields a meaningful performance increase, except in Aachen Day-Night. Introducing novel views into training improves performance on Aachen Day-Night, as well as on RobotCar Night. Surprisingly, however, its performance drops in CMU. 

\subsection{Additional Insights and Details}
\parsection{NeRF training details.}
In Sec. \ref{sec:supervision}, we detail our approach of dividing each scene into individual blocks along each trajectory and training a smaller NeRF for each block. Per trajectory, merging these smaller NeRFs results in a trajectory-NeRF. In our setup, however, we constructed a separate map for each trajectory pair, to generate more refined relative poses across trajectories. To remove scaling issues when comparing depth maps across trajectories from different maps, we train a trajectory-NeRF per trajectory for every trajectory pair. Furthermore, we expanded each block with buffer regions to improve the rendering quality of the NeRFs near the block edges. The buffer region is a 20\% extension of each block along each dimension. 

In Instant-ngp \cite{instant-ngp}, the representation space is divided into multiresolution hashed boxes, where each box is discretized into 128 steps along each axis. During ray-casting, there is one point sampled in each discrete step. Each box shares the same center, and are multiples of two larger than the smallest box, therefore the resolution of each discrete step in each box is a multiple of two smaller than in the smallest box. The purpose of this multiscale encoding is to speed up training, sampling points at a much higher rate near the center of the scene, however, at each iteration of training, the sampling point is randomized within each discrete step, therefore, even regions further from the center can be well represented with enough iterations. 

 \begin{figure}[t!]
 \centering
  \includegraphics[width=\linewidth]{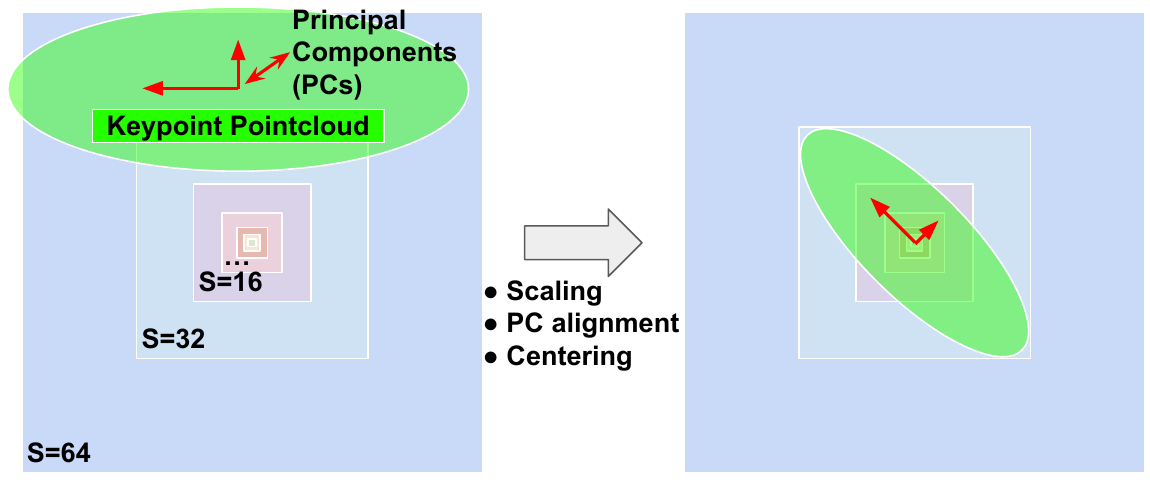}
\caption{\textbf{Optimized Bounding Box Alignment.} The keypoint pointcloud is the set of 3D points a map corresponding to a scene block. Initially, the pointclouds are scaled. Using Principal Component Analysis, we calculate the pointcloud's principal components. To fit as much of the scene into the smallest box possible, we center and transform the pointcloud to align the principal components with the body diagonal of the boxes.}
  \label{fig:box_centering}
\end{figure}

For best results, it is important to maximize the number of relevant scene points in the smallest bounding boxes. For each block, we scaled the scene such that the central bounding box (scale 1) spanned 3m of the scene, along all axes. Therefore, the largest bounding box (scale 64) spans 192m of the scene, defining the bounds of the NeRF representation. The scene surfaces in each block should be centered in the boxes. We aggregate the 3D points from the map corresponding to each frame in a block, and filter out points further than 64m from the corresponding camera frames. We center the scene around the median of the resulting pointcloud. Furthermore, we align the body diagonal of the boxes with the principal component of the pointcloud to further maximize the number of points in the smallest boxes.

\begin{figure*}[ht!]
 \centering
  \includegraphics[width=\linewidth]{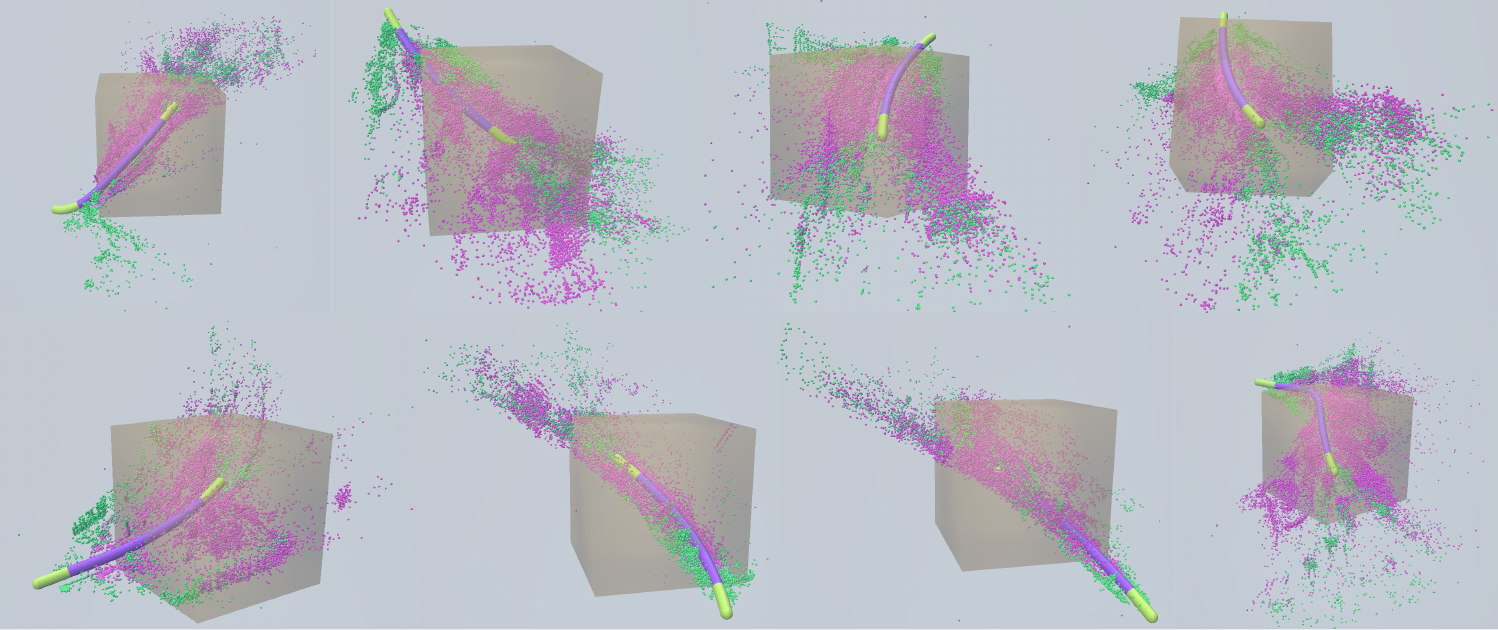}
\caption{\textbf{Bounding Box Alignment Illustrations.} Displayed are scenes aligned within scale 16 bounding boxes. Camera trajectories are depicted as tubes, with the pointcloud representing the 3D scene for each camera frame. Frames within blocks are shown as pink pointclouds and purple camera frames, while yellow camera frames and green pointclouds depict those in the buffer region.}
  \label{fig:blocks}
\end{figure*}

This process is illustrated in Fig. \ref{fig:box_centering}. Furthermore, in Fig. \ref{fig:blocks}, we present a collection of examples of the distributions of the pointclouds centered around the bounding boxes. Excluding the points corresponding to the buffer regions, approximately 80\% of the points are within the scale 16 bounding box.

Hyperparameters are selected per block to represent each block as accurately as possible. During training, certain training runs got stuck in local minima depending on initialization; the NeRF would estimate high density on the camera frame resulting in zero depth estimation. During a single training run, we reinitialize the training multiple times, each time transforming the scene slightly with respect to the bounding boxes. To evaluate the resulting trained NeRFs, we select the one that estimates depths that best fit the ground truths at corresponding keypoints. These ground truths are the extracted depths from our maps, which were used for depth supervision.
If no runs reached an average depth error lower than a certain threshold, all trained NeRFs were discarded. In that case, training was either restarted with new initialization, or abandoned after too many failed training runs. The abandoned blocks accounted for approximately 3\%. We train each NeRF for $3\cdot 10^5$ iterations. 

\parsection{Adapting R2D2 loss for cross-domain correspondences.}

To improve robustness to long-term localization performance, we use our generated correspondences to supervise feature extractors. We use R2D2 \cite{revaud2019r2d2} as the main network for our ablation study. We also trained SiLK \cite{gleize2023silk} to show generalization to other extractors. 

For an input image, R2D2 extracts pixel-wise descriptors and two score maps. The first score map estimates the repeatability of the corresponding points, while the second score map estimates the reliability of the corresponding descriptors.

To optimize for repeatability, a cosine similarity loss $\mathcal{L}_{cosim}$ and peakiness loss $\mathcal{L}_{peaky}$ are deployed. The repeatability loss $\mathcal{L}_{rep}$ is defined as the summation of the two losses:

\begin{equation}
    \label{eq:rep loss}
    \mathcal{L}_{rep} = \mathcal{L}_{peaky} + \mathcal{L}_{cosim}
\end{equation}

The peakiness loss enforces that the heatmap maintains a ``peaky" distribution, an essential addend in the repeatability loss function. For a more in-depth understanding, please refer to the original paper \cite{revaud2019r2d2}. The cosine similarity loss, on the other hand, penalizes the network, when the reference repeatability heatmap $\mathbf{S}$ and the query repeatability heatmap $\mathbf{S}'$ are dissimilar at corresponding pixel locations. In R2D2~\cite{revaud2019r2d2}, it is proposed to apply and average the cosine similarity on image patches, however, for our approach with filtered correspondences, we do not follow this patched approach. We instead use the following cosine similarity loss function:

\begin{equation}
    \label{eq:cosim loss}
    \mathcal{L}_{cosim}(\mathcal{C}) = 1 - cosim(\mathbf{S}[p_{1:N}], \mathbf{S}'[p'_{1:N}]),
\end{equation}
where $\mathcal{C}= \left\{ (p_i,p'_i) \right\}_{i\in N}$ is the set of $N$ pixel correspondences used for training. $\mathbf{S}[p_{1:N}]$ is a vector of heatmap values $\mathbf{S}$ at reference pixel locations, and $\mathbf{S}'[p'_{1:N}]$ is the vector of heatmap values $\mathbf{S}'$ at corresponding query pixel locations. To optimize for reliable descriptors, the descriptors and the reliability heatmap are jointly optimized, using the reliability loss, which is defined as:
\begin{equation}
    \label{eq:reliability loss}
    \mathcal{L}_{rel} = \frac{1}{N}\sum^N_{i=1}\mathcal{L}_{AP\kappa}(p_i),
\end{equation}
where
\begin{equation}
    \label{eq:AP loss}
    \mathcal{L}_{AP\kappa}(p) = 1-\left( AP(p)\mathbf{R}_p + \kappa(1-\mathbf{R}_p)\right).
\end{equation}
$\mathbf{R}_p$ is the estimated reliability value of pixel $p$, and $AP$ is a differentiable approximation of the Average Precision (AP) metric \cite{APmetric}. $1-AP(p)$ is then a ranking loss, which is minimized when the descriptor of $p$ is most similar to its positive anchor $p'$ and dissimilar to its negative anchors, randomly sampled from the set of points in the query image that maintain a specified minimum distance from $p'$. Eq. \ref{eq:AP loss} modifies this loss by introducing the repeatability estimation of the reference pixel. The purpose of optimizing the reliability and the descriptors together, is to estimate a low reliability if there is an anticipated low $AP(p)$ score.

The overall loss is the then the sum of the repeatability and the reliability loss terms:
\begin{equation}
    \label{eq:global loss}
    \mathcal{L}(\mathcal{P},\mathcal{P'}) = \mathcal{L}_{rep}(\mathcal{P},\mathcal{P'}) + \mathcal{L}_{rel}(\mathcal{P},\mathcal{P'})
\end{equation}

 We introduced iCDC to supervise extractors to generate domain invariant descriptors, and to extract keypoints that are useful for both cross- and intra-domain localization. While supervising with cross-domain correspondences are vital for extracting domain invariant descriptors, we propose to supervise keypoint detection with only intra-domain correspondences. Penalizing extracted keypoints that are not shared across a certain visual domain pair is not valid, because they might be useful keypoints across different domain pairs, or across intra-domain image pairs. 
Specifically, we turn off the repeatability loss in the cross-domain case and adapt the reliability loss. We adapt the overall loss function \ref{eq:global loss} to:
\begin{equation}
    \label{eq:adapted global loss}
    \mathcal{L}_{\mathcal{D},\mathcal{D'}}(\mathcal{P},\mathcal{P'}) = \mathbbm{1}_{\mathcal{D}=\mathcal{D'}}\cdot\mathcal{L}_{rep}(\mathcal{P},\mathcal{P'}) + \mathcal{L}_{rel,\mathcal{D},\mathcal{D'}}(\mathcal{P},\mathcal{P'}),
\end{equation}
where $\mathcal{D}$ is the viewing condition index of $\mathcal{P}$, and $\mathbbm{1}$ is the indicator function. 
$\mathcal{L}_{rel,\mathcal{D},\mathcal{D'}}$ is identical do $\mathcal{L}_{rel}$, except, in the case where $\mathcal{D}\neq\mathcal{D'}$, the gradients are not propagated through the reliability output of the network.

\begin{figure*}[t!]
    \hspace{-7pt}
    \centering
    \setlength\tabcolsep{1pt}
    \begin{tabular}{@{}r@{}}
   \includegraphics[width=0.84\linewidth]{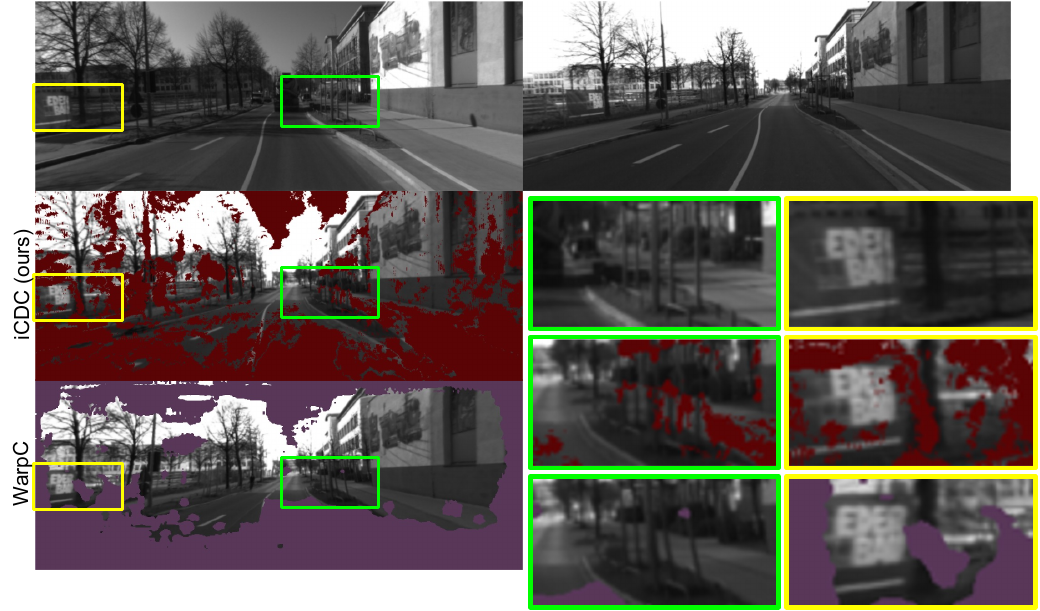} \\
  \end{tabular}
    \caption{\textbf{Additional Comparison Between Our NeRF Correspondences and WarpC~\cite{warpc} Correspondences.} Correspondences are generated for an image pair (top row). The left column's second and third rows show WarpC and iCDC correspondences respectively. More specifically, the right-hand side image is warped onto the left-hand side image following the correspondences. We zoom in on windows for detailed examination. In the case of accurate correspondences, the warped frames align with the left-hand side frame.}
    \label{fig:sup warpc vs nerf corresp}
\end{figure*}

\parsection{Additional results.}
In Sec. \ref{subsec:exp qualitative} of the main paper, we presented results comparing correspondences generated by a dense correspondence network (Warpc \cite{warpc}) and our iCDC method. In Fig. \ref{fig:sup warpc vs nerf corresp} we present an additional example, highlighting the improved performance of our correspondence generation method across domains. 

\begin{table*}[ht]
\setlength\tabcolsep{3pt}
  \center
    \caption{\textbf{Detailed CMU Seasons \cite{cmu} Benchmark Results}}

      \scriptsize
 \begin{tabular}{@{}rcccccccc@{}}\toprule

& Overcast & Sunny & Foliage & Mixed Foliage & No Foliage & Low Sun & Cloudy & Snowy\\ \midrule
SuperPoint~\cite{superpoint}  &  85.1 / 87.9 / 92.7 & 74.5 / 79.0 / 87.0 & 78.0 / 81.7 / 88.8 & 84.7 / 87.7 / 92.4 & 95.0 / 96.4 / 97.8 & 86.9 / 89.4 / 93.3 & 90.3 / 92.2 / 95.9 & 91.1 / 93.0 / 95.7 \\
R2D2~\cite{revaud2019r2d2}    &  87.7 / 90.9 / 94.2 & 77.2 / 82.1 / 87.5 & 80.7 / 85.0 / 89.6 & 85.2 / 89.0 / 92.9 & 94.8 / 96.6 / 97.9 & 86.8 / 90.2 / 93.6 & 91.6 / 94.1 / 96.2 & 90.8 / 93.5 / 95.8 \\
iCDC-R2D2                        &  \textbf{93.7} / \textbf{95.5} / \textbf{97.0} & \textbf{87.6} / \textbf{91.4} / \textbf{94.7} & \textbf{89.2} / \textbf{92.4} / \textbf{95.2} & \textbf{94.0} / \textbf{95.7} / \textbf{97.0} & \textbf{98.1} / \textbf{98.9} / \textbf{99.1} & \textbf{94.7} / \textbf{96.2} / \textbf{97.3} & \textbf{95.6} / \textbf{96.8} / \textbf{98.1} & \textbf{96.9} / \textbf{97.7} / \textbf{98.2} \\
\midrule
\midrule
SiLK \cite{gleize2023silk} & 39.3 / 43.8 / 52.3 & 33.6 / 37.8 / 44.0 & 35.9 / 40.1 / 47.1 & 36.3 / 40.7 / 49.4 & 56.1 / 60.3 / 67.8 & 42.5 / 46.6 / 54.8 & 44.8 / 49.4 / 57.8 & 43.1 / 47.3 / 56.4 \\
iCDC-SiLK & 80.4 / 83.3 / 87.1 & 72.0 / 76.5 / 81.2 & 74.4 / 78.2 / 82.6 & 80.1 / 83.3 / 87.4 & 91.3 / 93.3 / 95.1 & 83.0 / 85.8 / 89.4 & 84.4 / 86.9 / 89.6 & 85.5 / 88.2 / 91.0 \\
\bottomrule
\end{tabular}
\label{table:cmu benchmark}
\end{table*}

\begin{table*}[ht]
\setlength\tabcolsep{3pt}
  \center
    \caption{\textbf{Detailed RobotCar-Seasons \cite{robotcar} Benchmark Results}}

      \scriptsize
 \begin{tabular}{@{}rcccccccccc@{}}\toprule

& Night Rain & Overcast Winter & Sun & Rain & Snow & Dawn & Dusk & Night & Overcast Summer\\ \midrule
SuperPoint~\cite{superpoint}  &  12.8/23.6/27.1 & 57.3/97.0/\textbf{100} & \textbf{62.1}/87.9/\textbf{99.1} & 78.0/\textbf{94.6}/\textbf{100} & 68.8/97.7/\textbf{100} & 56.8/91.6/96.5 & 79.2/95.4/\textbf{100} & 15.0/31.4/39.8 & 47.9/92.9/98.6 \\ 
R2D2~\cite{revaud2019r2d2} & 16.3/30.0/33.5 & 58.5/\textbf{98.2}/\textbf{100} & 60.7/\textbf{89.3}/\textbf{99.1} & 78.0/\textbf{94.6}/\textbf{100} & 67.9/98.1/\textbf{100} & 57.3/91.6/96.5 & 79.7/\textbf{95.9}/\textbf{100} & 17.3/35.0/44.7 & \textbf{50.2}/\textbf{94.3}/\textbf{99.5} \\ 

iCDC-R2D2      &  \textbf{31.0}/\textbf{49.8}/\textbf{54.2} & \textbf{59.1}/97.6/\textbf{100} & 59.8/88.4/98.2 & 78.0/\textbf{94.6}/\textbf{100} & 68.8/98.1/\textbf{100} & 60.4/\textbf{95.2}/\textbf{100} & 79.7/95.9/\textbf{100} & \textbf{18.6}/\textbf{41.6}/\textbf{50.4} & 48.8/93.4/98.1 \\ 
\midrule
\midrule
SiLK \cite{gleize2023silk}    &  10.3/16.7/20.2 & 55.5/\textbf{98.2}/\textbf{100} & 47.3/70.1/84.4 & 77.6/\textbf{94.6}/\textbf{100} & \textbf{70.2}/97.2/99.1 & 60.8/93.4/98.7 & \textbf{80.2}/95.4/\textbf{100} & 14.6/19.9/23.0 & 44.5/89.6/92.4 \\ 
iCDC-SiLK    &  15.3/30.0/36.5 & 54.3/\textbf{98.2}/\textbf{100} & 58.9/87.1/98.2 & \textbf{78.5}/\textbf{94.6}/\textbf{100} & \textbf{70.2}/\textbf{98.6}/\textbf{100} & \textbf{61.2}/\textbf{95.2}/\textbf{100} & 79.7/95.4/\textbf{100} & 15.5/23.9/32.3 & 48.3/90.0/96.7 \\ 
\bottomrule
\end{tabular}
\label{table:robotcar benchmark}
\end{table*}

Furthermore, in Sec. \ref{subsec:exp longloc benchmark}, we presented our key results, evaluating our extractors on the long-term localization benchmark. In Tabs. \ref{table:cmu benchmark} and \ref{table:robotcar benchmark}, we present more detailed results for the same networks on the CMU Seaosns \cite{cmu} and RobotCar-Seasons benchmarks \cite{robotcar}. In CMU-Seasons, iCDC-R2D2 significantly outperforms other networks across all visual domains. On the other hand, the results are more balanced in RobotCar-Seasons. Similar to the less detailed results in the main paper, iCDC-R2D2 outperforms other extractors in the day-time scenario, however, not across all visual domains. In the main results, we found that the key improvements were in the night-time scenario. In these detailed results, we see that the greatest performance improvement occurs in the rainy night-time scenario.

\bibliographystyle{IEEEtran}
\bibliography{bibliography}

\clearpage
\begin{IEEEbiography}[{\includegraphics[width=1in,height=1.25in,clip,keepaspectratio]{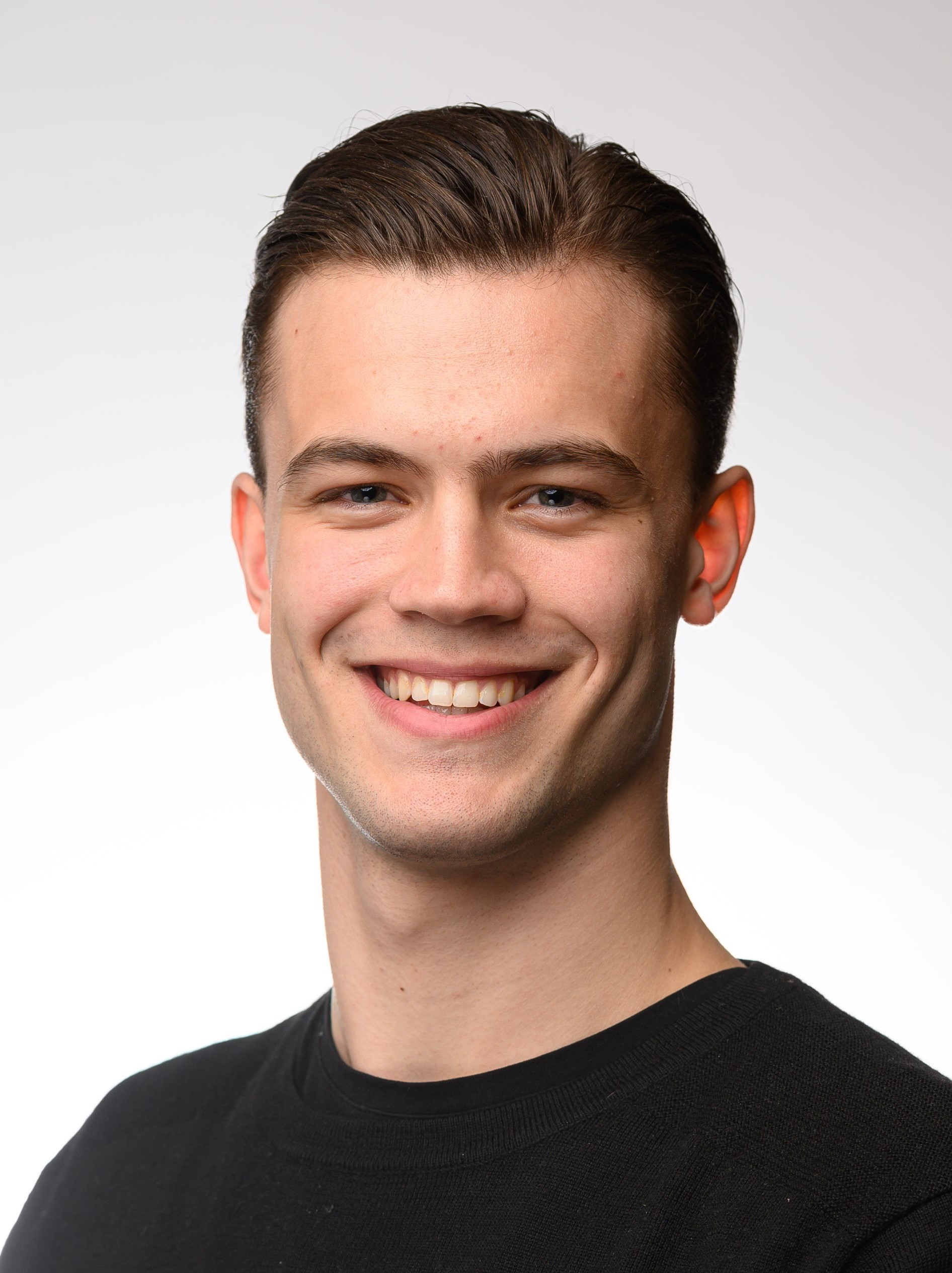}}]{Zador Pataki} recently completed his Master's degree and has now joined the Computer Vision and Geometry (CVG) lab at ETH Zurich to pursue a PhD. His research interests lie at the intersection of 3D Computer Vision and Deep Learning. During his Master's, he published papers in WACV and IROS. 
\end{IEEEbiography}

\begin{IEEEbiography}[{\includegraphics[width=1in,height=1.25in,clip,keepaspectratio]{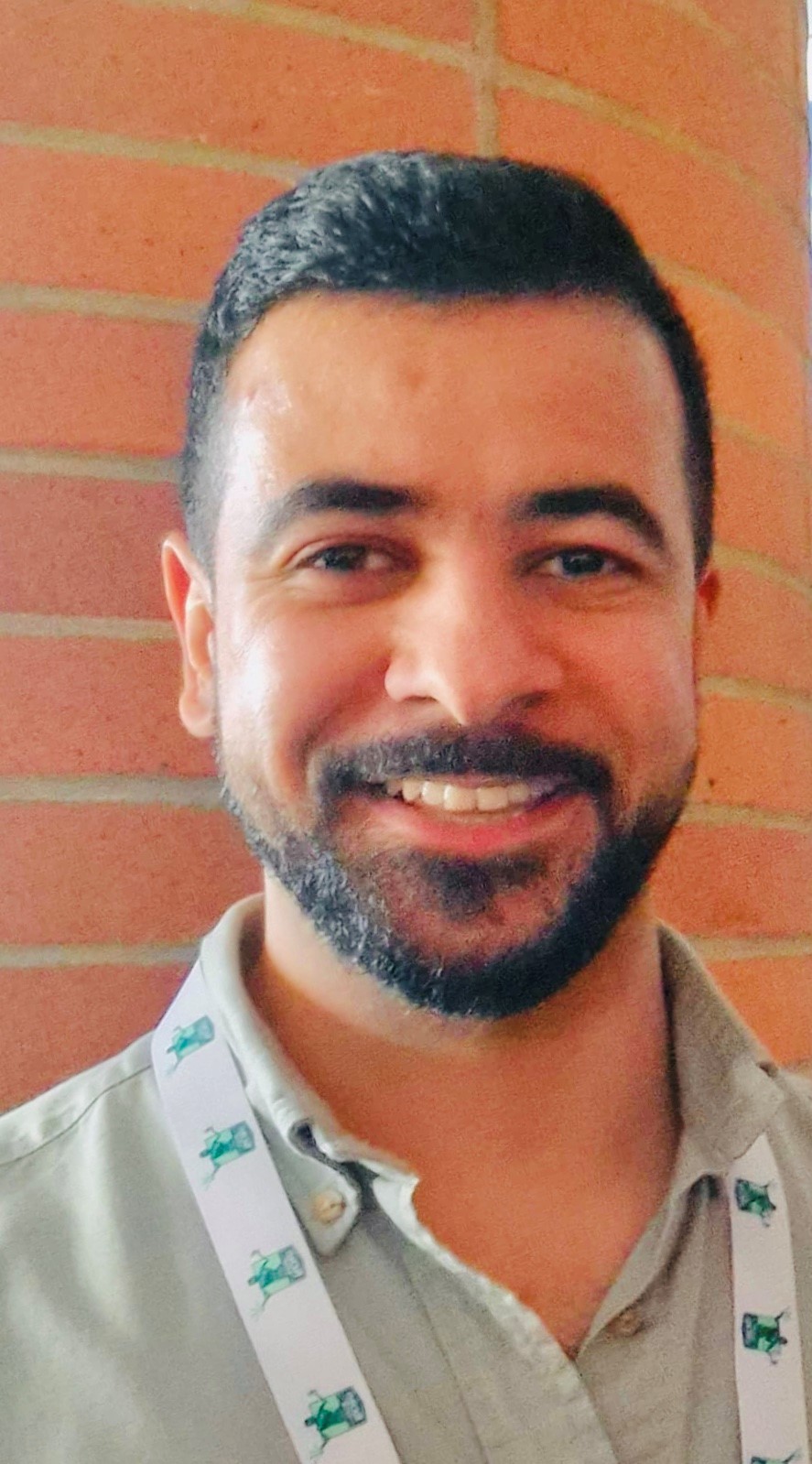}}]{Mohammad Altillawi}
is a PhD student at the Computer Vision Center (CVC) and the Autonomous University of Barcelona (UAB). He is conducting his research at Huawei Intelligent Robotics Cloud Computing lab in Munich. His research focuses on improving mapping and localization for robotics. Particularly, he aims to derive solutions that are accurate, real-time, and with low compute resources. Recently, he is focusing on leveraging novel view synthesis techniques towards improved mapping and localization. He has published in renown conferences, such as ICRA, IROS, and RA-L.
\end{IEEEbiography}

\begin{IEEEbiography}[{\includegraphics[width=1in,height=1.25in,clip,keepaspectratio]{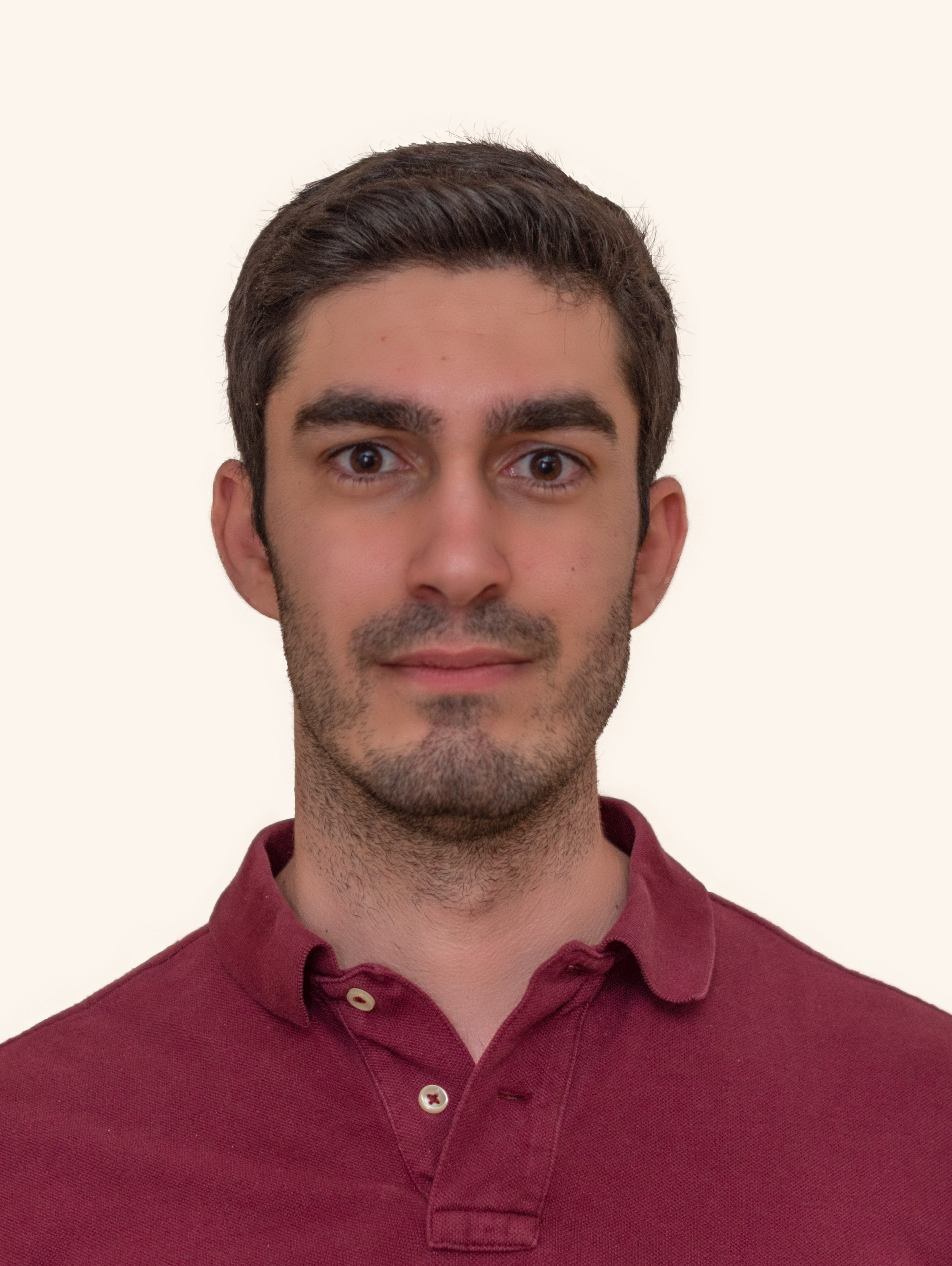}}]{Menelaos Kanakis}
earned his PhD from the Computer Vision Lab (CVL) at ETH Zurich. His research is focused on developing deep learning models under real-life constraints such as algorithmic run-time, limited annotations, or performing inference on-board computationally limited hardware. He has published in several renown conferences, such as CVPR, ICCV, ECCV, and ICML.

\end{IEEEbiography}

\begin{IEEEbiography}[{\includegraphics[width=1in,height=1.25in,clip,keepaspectratio]{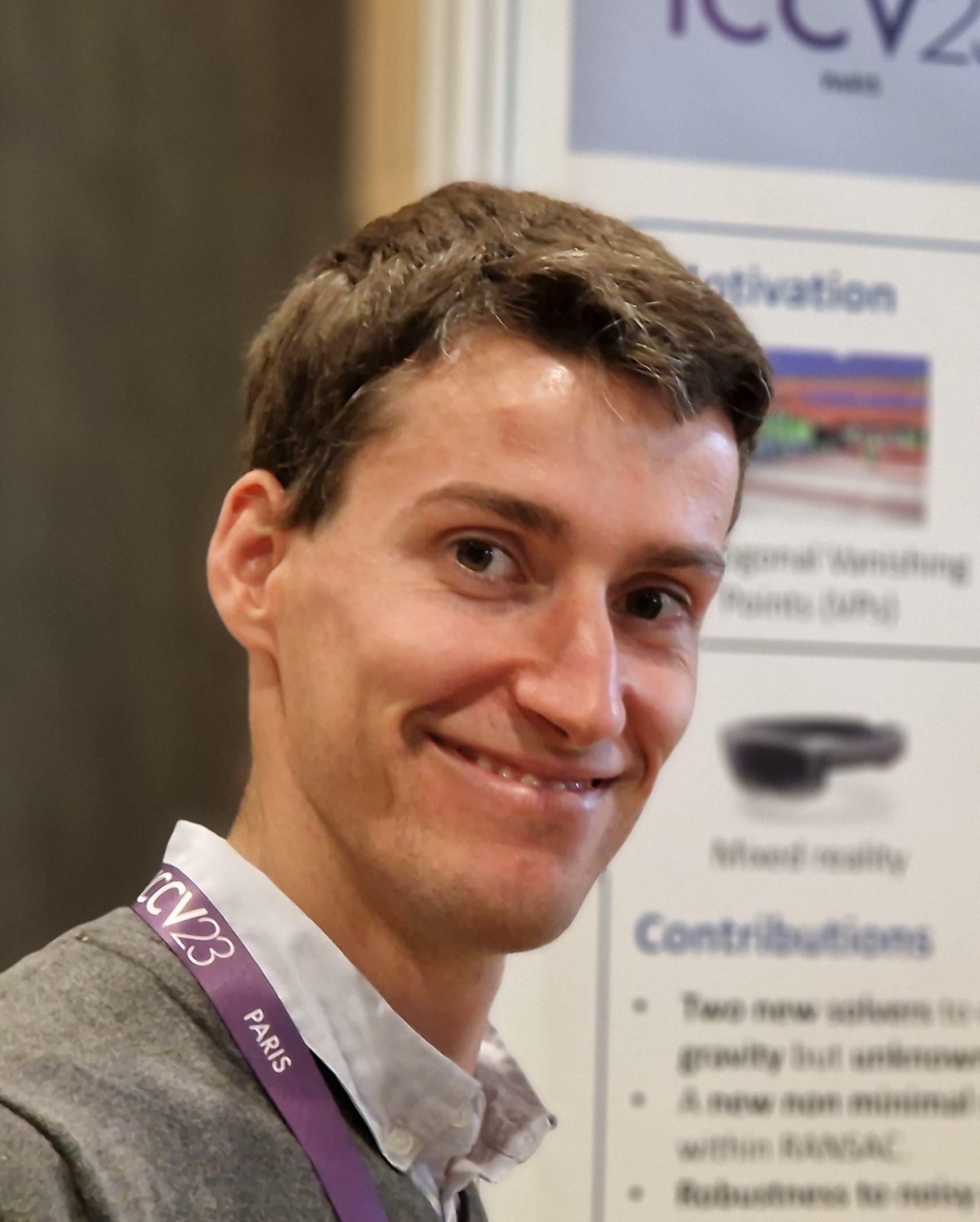}}]{Rémi Pautrat}
is a PhD student at the Computer Vision and Geometry (CVG) lab at ETH Zurich. His research interests are centered on local features and how to apply them to geometry tasks such as visual localization and 3D reconstruction. He focuses in particular on studying the invariance of local descriptors, and on higher-level features such as line segments, exploring ways to make the detection and matching pipelines as accurate and robust as possible. He has published in several renown conferences, such as CVPR, ICCV, ECCV, and ICRA.
\end{IEEEbiography}

\begin{IEEEbiography}[{\includegraphics[width=1in,height=1.25in,clip,keepaspectratio]{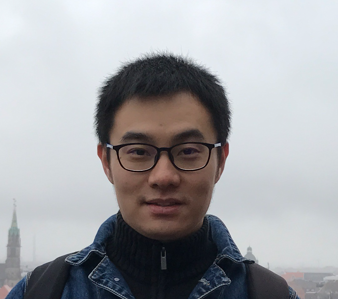}}]{Fengyi Shen}
is a PhD student at the Chair of Robotics, Artificial Intelligence and Embedded Systems at Technical University of Munich (TUM). He also works as a research intern in Huawei Munich Research Center and a collaborated research assistant in Embodied Perception and InteraCtion (EPIC) Lab at Peking Univerisity. His research interests include domain adaptive/generalizable and semi-supervised learning in computer vision, as well as generative models. He focuses in particular on exploring algorithms that can be trained in a label-efficient manner, aiming to reduce the effort of human annotation in computer vision tasks. He has published and served as reviewer in several renowned venues, such as CVPR, ECCV, BMVC, WACV and IEEE T-ITS.
\end{IEEEbiography}

\begin{IEEEbiography}[{\includegraphics[width=1in,height=1.25in,clip,keepaspectratio]{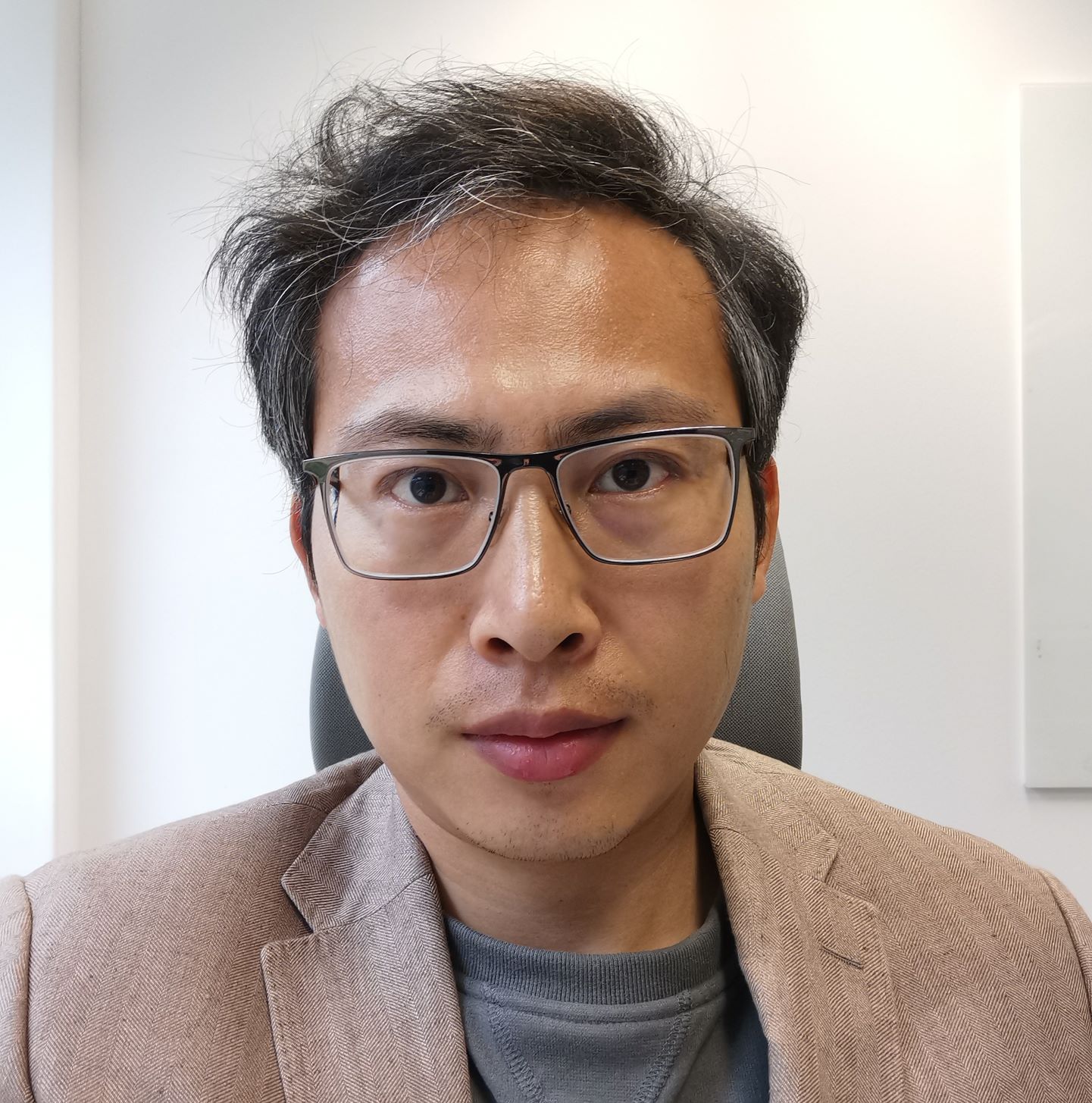}}]{Ziyuan Liu}
works as the chief scientist on AI robotics and simulation technologies at Huawei Cloud computing. Before joining Huawei, he was a tech lead on robotics at DOMO academy of Alibaba group. His research interests include simulation technolgoies and 3D robot vision. He has served as associate editor on IROS. In 2021 He launched the open cloud robot table top organization competition (OCRTOC) which is considered to be one of the first cloud-based robot competitions.
\end{IEEEbiography}

\begin{IEEEbiography}[{\includegraphics[width=1in,height=1.25in,clip,keepaspectratio]{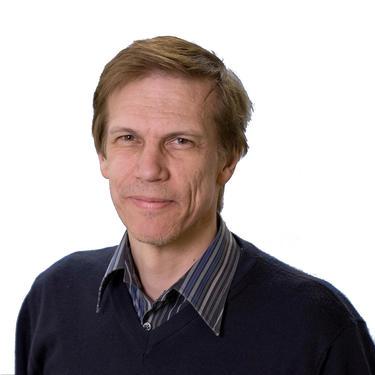}}]{Luc Van Gool}
is a full professor for Computer Vision at ETH Zurich and
the KU Leuven. He leads research and teaches at both places. He has authored over 300 papers. Luc Van Gool has been a program committee member of several, major
computer vision conferences (e.g. Program Chair ICCV'05, Beijing, General Chair of
ICCV'11, Barcelona, and of ECCV'14, Zurich). His main interests include 3D reconstruction and modeling, object recognition, and autonomous driving. He received several Best Paper awards (eg. David Marr Prize '98, Best Paper CVPR'07). He received the Koenderink Award in 2016 and `Distinguished Researcher' nomination by the IEEE Computer Society in 2017. In 2015 he also received they 5-yearly Excellence Prize by the Flemish Fund for Scientific Research. He was the holder of an ERC Advanced Grant (VarCity). Currently, he leads computer vision research for autonomous driving in the context of the Toyota TRACE labs in Leuven and at ETH, and has an extensive collaboration with Huawei on the issue of image and video enhancement.
\end{IEEEbiography}

\begin{IEEEbiography}[{\includegraphics[width=1in,height=1.25in,clip,keepaspectratio]{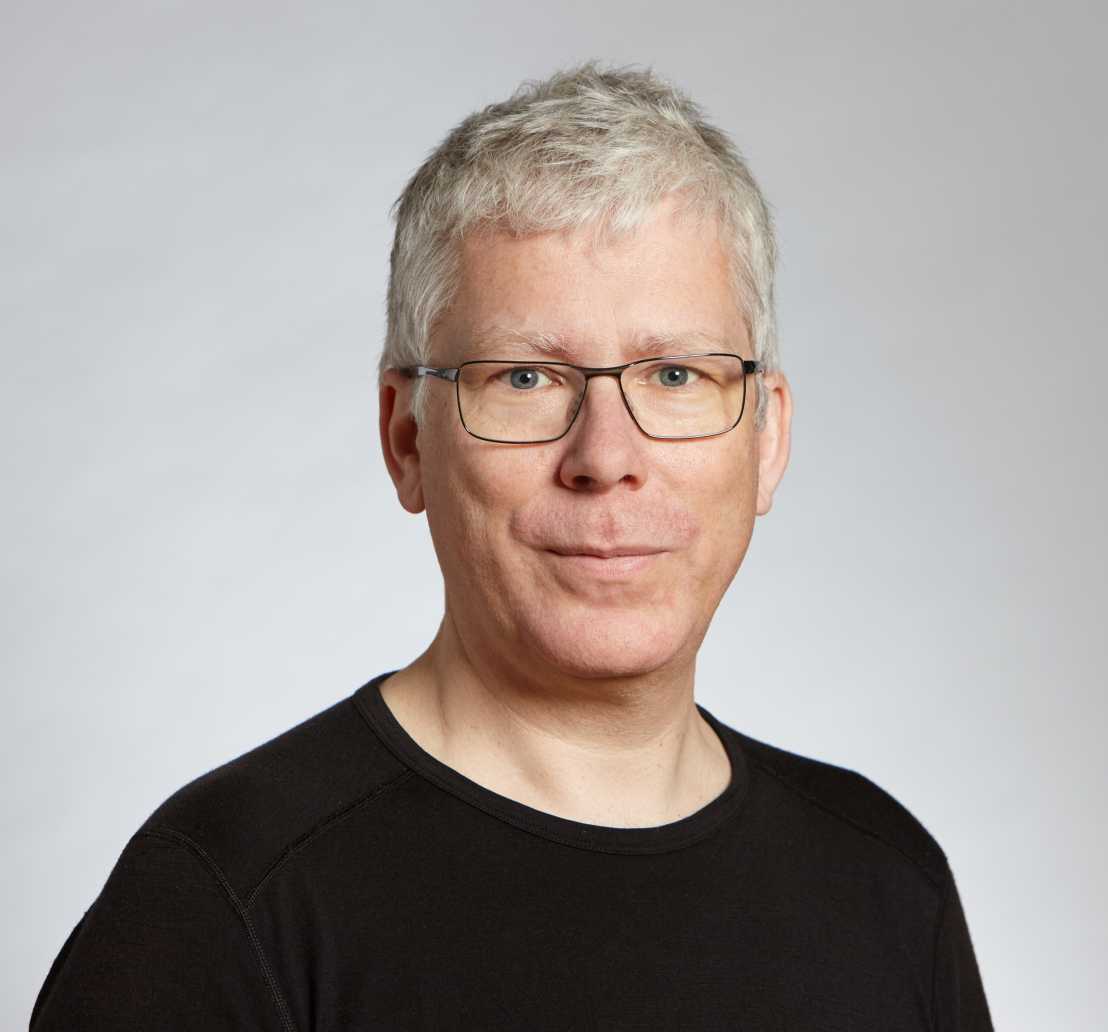}}]{Marc Pollefeys}
 is a full professor in the Dept. of Computer Science of ETH Zurich since 2007 where he leads the Computer Vision and Geometry lab. He is also the director of the Microsoft Mixed Reality and Artificial Intelligence Zurich lab where he leads a team of scientists working on perception algorithms for HoloLens and Mixed Reality. His main area of research is computer vision, but he is also active in robotics, machine learning and computer graphics. One of his main research goals is to develop flexible approaches to capture visual representations of real world objects, scenes and events. Prof. Dr. Pollefeys has received several prizes for his research, including a Marr prize, an NSF CAREER award, a Packard Fellowship and a European Research Council Starting Grant. He is the author or co-author of more than 300 peer-reviewed publications. He was the General Chair of ICCV 2019 and ECCV 2014 and Program Co-Chair for CVPR 2009. Prof. Pollefeys has served on the Editorial Board of the IEEE Transactions on Pattern Analysis and Machine Intelligence, the International Journal of Computer Vision and Foundations and Trends in Computer Graphics and Computer Vision. He is a fellow of the IEEE.
\end{IEEEbiography}

\end{document}